\documentclass[conference]{IEEEtran}
\IEEEoverridecommandlockouts
\usepackage{cite}
\usepackage{epsfig}
\usepackage{amsmath,amssymb,amsfonts}
\usepackage{algorithmic}
\usepackage{graphicx}
\usepackage{textcomp}
\usepackage{xcolor}
\usepackage{subcaption}
\usepackage{multirow}
\usepackage{caption}
\usepackage[ruled,vlined]{algorithm2e}
\usepackage{mathtools}
\DeclarePairedDelimiter\floor{\lfloor}{\rfloor}

\SetKwInput{KwInput}{Input}                
\SetKwInput{KwOutput}{Output}              

\def\BibTeX{{\rm B\kern-.05em{\sc i\kern-.025em b}\kern-.08em
    T\kern-.1667em\lower.7ex\hbox{E}\kern-.125emX}}
\begin{document}

\title{Exploring the Back Alleys: Analysing The Robustness of Alternative Neural Network Architectures against Adversarial Attacks\\
}

\author{\IEEEauthorblockN{Yi Xiang Marcus Tan,\IEEEauthorrefmark{1}\IEEEauthorrefmark{3} Yuval Elovici,\IEEEauthorrefmark{1}\IEEEauthorrefmark{2}, and Alexander Binder\IEEEauthorrefmark{1}\IEEEauthorrefmark{3}}\\
	\IEEEauthorblockA{\IEEEauthorrefmark{1}ST Engineering Electronics-SUTD Cyber Security Laboratory\\
		\IEEEauthorrefmark{3}Information Systems Technology and Design (ISTD) Pillar,
		Singapore University of Technology and Design\\
		\IEEEauthorrefmark{2}Department of Software and Information Systems Engineering, Ben-Gurion University of the Negev\\
		\IEEEauthorrefmark{2}Deutsche Telekom Innovation Laboratories at Ben-Gurion University of the Negev
	}}

\maketitle


\begin{abstract}
We investigate to what extent alternative variants of Artificial Neural Networks (ANNs) are susceptible to adversarial attacks. We analyse the adversarial robustness of conventional, stochastic ANNs and Spiking Neural Networks (SNNs) in the raw image space, across three different datasets. 
Our experiments reveal that stochastic ANN variants are almost equally as susceptible as conventional ANNs when faced with simple iterative gradient-based attacks in the white-box setting. 
However we observe, that in black-box settings, stochastic ANNs are more robust than conventional ANNs, when faced with boundary attacks, transferability and surrogate attacks. Consequently, we propose improved attacks and defence mechanisms for stochastic ANNs in black-box settings.
When performing surrogate-based black-box attacks, one can employ stochastic models as surrogates to observe higher attack success on both stochastic and deterministic targets. This success can be further improved with our proposed Variance Mimicking (VM) surrogate training method, against stochastic targets.
Finally, adopting a defender's perspective, we investigate the plausibility of employing stochastic switching of model mixtures as a viable hardening mechanism. We observe that such a scheme does provide a partial hardening.

\end{abstract}

\section{Introduction}

Second generation neural networks have been empirically successful in solving a plethora of tasks. Different variants of Artificial Neural Networks (ANN) have been used in image recognition \cite{krizhevsky2012imagenet} in all kind of forms, natural language processing, detecting anomalous behaviours in cyber-physical systems \cite{7911887}, or simply playing a game of Go \cite{silver2016mastering}. 

In 2013, first research showcased the vulnerability of ANNs to adversarial attacks \cite{szegedy2013intriguing},
a phenomenon that involves the creation of perturbed samples from their original counterparts, imperceptible upon visual inspection, which are misclassified by ANNs. Since then, many researchers introduced other adversarial attack methods against such ANN models, whether under a white-box \cite{Goodfellow2014,madry2017towards,kurakin2016adversarial,moosavi2016deepfool,Carlini2017} or a black-box
\cite{Brendel2018,papernot2017practical} scenarios. This raises questions about the reliability of ANNs, which can be a cause for concern especially when used in cyber-security or mission critical contexts. \cite{tan2019adversarial,Rosenberg2018}. 

While not reaching state of the art accuracies, Spiking Neural Networks (SNNs), are investigated as a means to model the biological properties of the human brain more closely as compared to their ANN counterparts. In contrast to ANNs, SNNs train on spike trains rather than image pixels or a set of predefined features. There have been different variants of SNNs, differing in terms of the learning rule used (whether through standard backpropagation \cite{Lee2019,sengupta2019going,Huh2018} or via Spike-Timing-Dependent Plasticity (STDP) \cite{Diehl2015,kheradpisheh2018stdp,Jeyasothy2019}) or the architecture. In this work, we focused on the STDP-based learning variant of SNNs.
Stochastic ANNs have also been used to perform image classification tasks.
In this work, we focused on two sub-categories of such stochastic ANNs, one involving making both its hidden weights and activations are in a binary state \cite{Hubara2016}, while the other only requiring its hidden activations to be binary \cite{Bengio2013,Raiko2014,Yin2019}. These variants of networks use Bernoulli distributions in order to binarize its features. 
Since there are strong evidences showcasing the weaknesses of ANNs to adversarial attacks, we question if there exists alternative variants of neural networks that are inherently less susceptible to such a phenomenon. 


The authors in \cite{Sharmin2019} gave a preliminary study of investigating the adversarial robustness of two variants of SNNs that used gradient backpropagation during training, namely ANN-to-SNN conversion \cite{sengupta2019going} and also Spike-based training \cite{Lee2019}. The authors examined the robustness of the SNNs, and also a VGG-9 model in the white-box and black-box settings. They concluded that SNNs trained directly on spike trains are more robust to adversarial attacks as compared to SNNs converted from their ANN counterparts.
However, in their experiments, the authors performed their attacks on intermediate spike representations of images, which is the result of passing images through a Poisson Spike Generation phase followed by rate computation. Though their work shows preliminary results on the robustness of SNNs, we find that their simplified approach of constructing adversarial samples yields unrelatable deviations between the natural and their adversarial counterparts in the image space. 
We attempt to address those points in our work, by focusing on STDP-based learning SNNs and also constructing adversarial samples in the input space. \cite{Galloway2017,Khalil2018} explored adversarial attacks against Binary Neural Networks (BNNs).
To the best of our knowledge, we did not find prior work examining the adversarial robustness of networks employing the use of Binary Stochastic Networks (BSN). The authors in \cite{Galloway2017} performed two white-box attacks and a black-box attack (the Fast Gradient Sign Method (FGSM) \cite{Goodfellow2014}, CWL2 and the transferability from a deterministic substitute model proposed by \cite{papernot2017practical}) and showed that stochasticity in binary models does improve the robustness against attacks.


Unlike \cite{Sharmin2019}, we examine two very recent works in the field of SNN: the Multi-Class Synaptic Efficacy Function-based leaky-integrate-and fire neuRON (MCSEFRON) model \cite{Jeyasothy2019} and Reward-modulated STDP spike-timing-dependent plasticity in deep convolutional network proposed in \cite{Mozafari2019}. We refer to the latter model as $SNN_{m}$ for notational simplicity. For our stochastic ANN variants, we use Binary Stochastic Nets (BSN) to give our models binarized activations in a stochastic manner. Also, we used Binarized Neural Networks (BNN) that binarizes weights and activations as our second variant of the stochastic ANN. We used the vanilla ResNet18 model as a bridge across the different variants of neural networks. The contributions of our work are as follows:
\begin{enumerate}
    \item We analyse to what extent adversarial attacks (white-box and black-box) can be performed in the original image space against SNNs with different information encoding schemes. We analyze the effectiveness of adversarial attacks against stochastic neural network models. We employ the vanilla ResNet18 CNN as a baseline for comparison with above models.
    
    
    
    
    \item For white box settings, we propose, inspired by \cite{eot}, an alternative objective function for the state-of-the-art CWL2 white-box attack and we analyse the robustness of the different network variants to samples generated via such attacks. 
    
    \item For black-box setups, which compared to the white-box ones, make more realistic assumptions about the information obtainable about a model hidden behind a service, we investigate the susceptibility of alternative variants of neural networks against boundary attacks, directly transferred adversarial samples across architectures, and surrogate-based attacks.
    
    \item Based on our observation of increased adversarial robustness of stochastic ANNs in black-box settings, we propose a stronger surrogate-based attack. This is based on using stochastic ANNs as surrogates, and as a novel contribution, Variance Mimicking. 
    
    \item As a second novel contribution for the black-box setting, we propose stochastic mixtures of networks with different architectures as hardening measure. We measure the efficiency of attacks against stochastic mixtures of different architectures. Given the availability of different variants of neural networks, a stochastic mixture of them is an feasible hardening mechanism which improves robustness.
\end{enumerate}
We present our work from the perspective, that white-box settings are amenable to theoretical analysis, while black-box scenarios are more likely to be encountered in practice. 
From a practitioner's perspective, a service provided to a customer will not expose the model internals. 
For that reason in the second part, we focus on black-box setups, assuming that the attacker has access to only the model predictions. Secondly, we assume that perfect defence is not achievable, same as perfect prediction accuracies are not. Therefore we consider hardening, to make attacks more costly, and we investigate how stochastic ANNs and stochasticity in general can be employed for that purpose. Costlier attacks can be detected by other means, namely as high volume requests consisting of very similar data points. This justifies hardening as research effort, when perfect adversarial detection is not feasible. By this work, beyond mere technical aspects, we invite the research community to reconsider the options for defences against adversarial attacks.
The remaining of our paper is organised as such: 
we present a brief introduction to SNNs and stochastic ANNs and attack details we used in Section \ref{background}. In Section \ref{exp}, we discuss our experiments and findings. 
This is followed by a discussion of a plausible hardening mechanism by using stochastic architecture mixtures in Section~\ref{stocswitch}.
After which, in Section~\ref{discuss}, we provide deeper analysis with regards to stochastic ANNs and conclude our work in Section \ref{conclude}.

\section{Background}
\label{background}

\subsection{Adversarial Attacks Against Neural Networks}
The concept of adversarial examples were first introduced by \cite{szegedy2013intriguing}.
Following their work, several other researchers explored various methods to launch adversarial attacks in an attempt to further evaluate the robustness of ANNs. One variant is FGSM, which uses the sign of the gradients computed from the loss, to perform a single-step perturbations on the input itself. 
Several studies \cite{madry2017towards,kurakin2016adversarial} 
extended this technique by applying the algorithm to the input image sample for multiple iterations to construct a stronger adversarial samples. The CWL2 attack \cite{Carlini2017} is one of the state-of-the-art white-box adversarial attack method, capable of producing visually imperceptible, yet misclassified images, that are robust against \textit{defensive distillation} \cite{papernot2016distillation}.

The methods described above and many other methods proposed by the scientific community \cite{moosavi2016deepfool,papernot2016limitations,modas2019sparsefool,Liu2016} pertain to attacks done in a white-box setting, in which it is assumed that the attacker has full knowledge and access to the ANN image classifier. However, several researchers \cite{Brendel2018,papernot2017practical} 
have also shown that it is also possible to attack a model without knowledge of the targeted model (i.e. black-box attacks). In \cite{Brendel2018}, the authors used the decision made by the targeted image classifier to perturb the input sample. In \cite{papernot2017practical}, 
the authors made use of the concept of transferability of adversarial samples across neural networks to attack the victim classifier, by approximating the decision boundary of the targeted classifier by training a surrogate model. 
In the next section, we describe the attacks we used in our work, exploring both the white-box and black-box categories.

\subsection{Attack Algorithms Used}
To attack the model in a black-box setting, we used a decision-based method known as \textit{Boundary Attack} \cite{Brendel2018}.
This approach initialises itself by generating a starting sample that is labelled as adversarial to the victim classifier. Then, random walks are taken by this sample along the decision boundary that separates the correct and incorrect classification regions. These random walks will only be considered valid if it fulfils two constraints, i) the resultant sample remains adversarial and ii) the distance between the resultant sample and the target is reduced. Essentially, this approach performs rejection sampling such that it finds smaller valid adversarial perturbations across the iterations.

We used the \textit{Basic Iterative Method} (BIM) \cite{kurakin2016adversarial} as one of the means to perform white-box attacks. This method is basically an iterative form of the FGSM: 
\begin{align}
    x_{t+1} &= x_t + \alpha * sign(\nabla J(F(x_t), y; \theta))
\end{align}
\noindent where $\nabla J$ represents the gradients of the loss calculated with respect to the input space $x_t$ and its original label $y$, $t$ represents the iterations. 

The second white box attack used is the CWL2 attack. It is based on solving the following objective function:
\begin{align}
    &min_\delta ||\delta||_2 + c\cdot f(x+\delta)
    \label{cweqn}
\end{align}
\noindent where the first term minimises the $L_2$ norm of the perturbation while the second term ensures misclassification. $c$ is a constant.
This attack method is considered as state-of-the-art and can bypass several detection mechanisms \cite{Carlini2017bypass}.

\subsection{Spiking Neural Networks}

MCSEFRON \cite{Jeyasothy2019} is a two-layered SNN that has time-dependent weights connecting between neurons. It adopts the STDP learning rule and it trains based on variations between the relative timings between the actual and desired post-synaptic spike time. It encodes images into spike trains via the same mechanism as \cite{bohte2002error}, which involves projecting the real-valued normalised image pixels (in [0,1]) onto multiple overlapping receptive fields (RF) represented by Gaussians. After the training is done, it makes decisions based on the earliest post-synaptic spikes while ignoring the rest. 

SNN$_M$ \cite{Mozafari2019} is an architecture which uses three convolution layers, with the first two trained in an unsupervised manner via STDP and the last convolution trained via Reward-modulated STDP. The input images had to be first preprocessed by six Difference of Gaussian (DoG) filters, which were followed by the encoding into spike trains by the intensity-to-latency \cite{gautrais1998rate} scheme. The SNN$_M$ does not require any external classifiers as they used a neuron-based decision-making trained via R-STDP in the final convolution layer. The R-STDP is based on reinforcement learning concepts, where correct decisions will lead to STDP while incorrect decisions will lead to anti-STDP.

\section{Experiments and Results}
\label{exp}

We used three datasets for our experiments, namely MNIST \cite{mnist}, CIFAR-10 \cite{CIFAR10} and, Patch Camelyon \cite{Veeling2018qh} which we refer to as PCam. The libraries we used in our experiments are PyTorch \cite{paszke2017automatic} and SpykeTorch \cite{spyketorch} for constructing our image classifiers. For attacks, we used the Foolbox \cite{rauber2017foolbox} library at version 1.8.0.

\subsection{Image Classification Baseline}

In this work, we explored eight different variants of neural networks: ResNet18, MCSEFRON, SNN$_M$, three BSN architectures and two BNN architectures. The BSN architectures used are a 2-layered, 4-layered Multilayer Perceptron, and a modified LeNet \cite{mnist} which we will refer to as BSN-2, BSN-4 and BSN-L respectively. For the BNNs, we explored both deterministic and stochastic binarization strategies, which we will refer to as BNN-D and BNN-S respectively. We leave details of training our classifiers to the Appendix \ref{trainingclfs}.

\subsubsection{Baseline Classification Performance}

The baseline image classification accuracies are summarised in Table~\ref{baseline}. It is clear that these are not state-of-the-art. Getting the best performance is not the focus of this work since we are interested in adversarial robustness. 
As an example, MCSEFRON shows poor performance on CIFAR-10. It can be considered as a single layered neural network without any convolution layers. 
In a prior work that studied the performance limitations of models without convolutions \cite{lin2015far}, they managed to obtain an accuracy of only approximately 52\% to 57\% on CIFAR-10, using a deeper and more dense fully-connected neural network (see Figure 4(a) in \cite{lin2015far}).
\begin{table}[tb]
\centering
\caption{Baseline image classification performance for all models}
\label{baseline}
\resizebox{0.4\textwidth}{!}{%
\begin{tabular}{cccc}
\hline
         & MNIST & CIFAR-10 & PCam  \\ \hline
Resnet18 & 0.988 & 0.842    & 0.789 \\ \hline
MCSEFRON & 0.861 & 0.372    & 0.671 \\ \hline
SNN$_M$  & 0.964 & 0.391    & -     \\ \hline
BSN-2    & 0.962 & 0.488    & 0.730 \\ \hline
BSN-4    & 0.972 & 0.542    & 0.733 \\ \hline
BSN-L    & 0.990 & 0.642    & 0.788 \\ \hline
BNN-D    & 0.989 & 0.876    & 0.798 \\ \hline
BNN-S    & 0.967 & 0.647    & 0.780 \\ \hline
\end{tabular}
}
\end{table}
\subsection{White-box Attacks Against Neural Networks}
We report the proportion of adversarial samples that are successful in causing misclassification and term it as Adversarial Success Rate (ASR; in range [0,1]). Furthermore, we report the mean $L_2$ norms per pixel, of the differences between natural images and their adversarial counterparts. 
In our experiments, we sub-sampled 500 samples from the test set of the respective datasets during the evaluation of the BIM attack and 100 samples for the evaluation of the other attacks, unless stated otherwise. We performed sub-sampling due to the computational intractability of performing the attacks on the entire dataset. Note that we selected only samples that were \textit{originally correctly classified}. For stochastic ANNs, we adopted an average inference policy, taking the average prediction across 10 forwards passes per sample. We refer the readers to Appendices~\ref{evalstoc} and \ref{modsnn} for more experimental details on evaluation of stochastic ANNs and SNNs due to space constraints.

\subsubsection{Basic Iterative Method (BIM)}
For the BIM attack, we varied the attack strength (symbolised by $\epsilon$ measured in $L_\infty$ space) while keeping the step sizes and iterations fixed at 0.05 and 100 respectively. We explored $\epsilon$ values of $8/255$, $16/255$ and $32/255$ in our experiments, showing the results of $\epsilon=32/255$ while the rest can be found in Appendix~\ref{fullbim}.
\begin{table*}[htb]
\centering
\label{otherattackmain:otherattackmain}
\caption{Adversarial success rate (Table (a)) and mean $L_2$ norms per pixel (Table (b)) for the attacks. 
For ModCWL2, $K$ in Equation \ref{modcweqn} was set to 5}
\begin{subtable}[h]{\textwidth}
\centering
\caption{ASR (in {[}0,1{]}) of the different variants of models.}
\label{otherattackmain:otherattacks}
\resizebox{0.87\textwidth}{!}{%
\begin{tabular}{lccccccccc}
\hline
Dataset                   & \multicolumn{1}{l}{Attack Method} & Resnet18 & SNN$_M$ & MCSEFRON & BSN-2  & BSN-4  & BSN-L   & BNN-D & BNN-S  \\ \hline
\multirow{4}{*}{MNIST}    & BIM                               & 1.000    & 0.120   & 0.294    & 0.8741 & 0.9556 & 0.8306  & 1.000 & 0.5949 \\
                          & CWL2                              & 0.970    & 0.620   & 0.420    & 0.8126 & 0.8487 & 0.7772  & 0.980 & 0.2998 \\
                          & ModCWL2                           & 1.000    & 1.000   & 1.000    & 0.8075 & 0.8656 & 0.7853  & 1.000 & 0.2934 \\
                          & Boundary                          & 1.000    & 1.000   & 1.000    & 0.1577 & 0.0441 & 0.0026  & 0.980 & 0.0443 \\ \hline
\multirow{4}{*}{CIFAR-10} & BIM                               & 1.000    & 0.694   & 0.998    & 0.9874 & 0.9880 & 0.9861  & 1.000 & 0.9983 \\
                          & CWL2                              & 1.000    & 0.990   & 0.990    & 0.582  & 0.7973 & 0.7892  & 1.000 & 0.6603 \\
                          & ModCWL2                           & 1.000    & 1.000   & 1.000    & 0.9402 & 0.8545 & 0.8509  & 1.000 & 0.6926 \\
                          & Boundary                          & 1.000    & 1.000   & 1.000    & 0.7417 & 0.4660 & 0.4197  & 0.944 & 0.4448 \\ \hline
\multirow{4}{*}{PCam}     & BIM                               & 1.000    & -       & 0.534    & 0.9870 & 0.9549 & 0.91544 & 0.974 & 0.9739 \\
                          & CWL2                              & 1.000    & -       & 0.280    & 0.4932 & 0.4755 & 0.4772  & 0.9200  & 0.3216 \\
                          & ModCWL2                           & 1.000    & -       & 0.800    & 0.6024 & 0.5471 & 0.4627  & 0.930 & 0.3956 \\
                          & Boundary                          & 0.730    & -       & 1.000    & 0.0117 & 0.005  & 0.0151  & 0.190 & 0.0081 \\ \hline
\end{tabular}
}
\end{subtable}
\begin{subtable}[h]{\textwidth}
\centering
\caption{Mean $L_2$ norms per pixel (upscaled by 1000 for illustration) between the original image and its perturbed adversarial image of the different variants of models.}
\label{otherattackmain:otherattackl2norm}
\resizebox{0.87\textwidth}{!}{%
\begin{tabular}{lccccccccc}
\hline
Dataset                   & Attack Method & Resnet18 & SNN$_M$ & MCSEFRON & BSN-2   & BSN-4  & BSN-L  & BNN-D  & BNN-S  \\ \hline
\multirow{4}{*}{MNIST}    & BIM           & 2.1667   & 2.4142  & 1.4126   & 2.5725  & 2.2894 & 2.6324 & 1.5606 & 2.2744 \\
                          & CWL2          & 0.9057   & 3.5731  & 0.5137   & 2.1460  & 2.4674 & 2.5942 & 0.0000 & 1.4761 \\
                          & ModCWL2       & 5.8529   & 7.7747  & 4.6039   & 2.2873  & 2.4644 & 2.6049 & 0.2909 & 1.7559 \\
                          & Boundary      & 1.3986   & 10.7922 & 3.4964   & 7.7322  & 5.4017 & 5.0642 & 0.0000 & 0.5951 \\ \hline
\multirow{4}{*}{CIFAR-10} & BIM           & 0.9318   & 1.3968  & 0.8924   & 1.1485  & 1.0247 & 1.0947 & 0.9606 & 1.0407 \\
                          & CWL2          & 0.0782   & 0.5601  & 0.0724   & 0.0862  & 0.0867 & 0.0916 & 0.0376 & 0.0878 \\
                          & ModCWL2       & 0.1102   & 0.4354  & 0.0766   & 5.9525  & 0.1908 & 0.2053 & 0.0640 & 0.1048 \\
                          & Boundary      & 0.1346   & 2.3423  & 2.2432   & 3.6948  & 2.9595 & 1.8603 & 1.1771 & 2.3559 \\ \hline
\multirow{4}{*}{PCam}     & BIM           & 0.9794   & -       & 1.4248   & 0.9491  & 0.9426 & 1.0308 & 1.0343 & 0.9882 \\
                          & CWL2          & 0.0870   & -       & 0.7915   & 0.5155  & 0.8428 & 1.0637 & 0.1001 & 0.5285 \\
                          & ModCWL2       & 0.1367   & -       & 3.2671   & 1.0358  & 0.8888 & 1.0759 & 0.1384 & 0.5325 \\
                          & Boundary      & 0.0856   & -       & 3.1918   & 1.66239 & 1.3100 & 2.3357 & 2.0002 & 2.4087 \\ \hline 
\end{tabular}}
\end{subtable}

\end{table*}

Two notable observations can be made about BIM from Tables \ref{otherattackmain:otherattacks} and \ref{otherattackmain:otherattackl2norm}:
Firstly, when comparing networks, BIM has the highest ASR against BNN-D other than ResNet18. We highly suspect that this is a result of the binarization policy that was used in this variant. As the binarization of hidden weights and activations in the forward pass is a sign function, it becomes easier to cause label flipping as the magnitude is no longer as significant, so long as the sign can be flipped.
Secondly, when comparing attacks for a given architecture, BIM yields the highest ASR on BNN-D and BNN-S, however this is achieved at the cost of higher L2-norms. We defer an explanation for this phenomenon to Section~\ref{stocsuscep}.

\subsubsection{Carlini \& Wagner L2 (CWL2)}
\label{cwmethod}
For the CWL2 attack, we used the default attack parameters as specified in Foolbox. For stochastic ANNs, we disabled binary searching for the constant $c$ and showing the results when $c=10$ in Tables~\ref{otherattackmain:otherattacks} and \ref{otherattackmain:otherattackl2norm}. For deterministic variants, binary search was enabled. Exemplified by the results from ResNet18 in Table~\ref{otherattackmain:otherattacks}, the CWL2 attack is an extremely powerful attack that manages to compromise the model almost all of the time.
However, this attack is less effective against stochastic ANNs compared to BIM, evident in Table~\ref{otherattackmain:otherattacks} where BIM outperforms CWL2 in all stochastic ANN settings.
Although this attack method is known to be state-of-the-art in generating successful adversarial samples with the least perturbation, its efficacy drops significantly when faced with stochastic model variants.
This observation, together with the inspiration from a prior art in \cite{eot}, prompted us to propose a modification of the objective function of this attack, which will be discussed next.

\subsubsection{Augmented Carlini \& Wagner L2 Attack Against Neural Networks (ModCWL2)}
\label{augcwattack}
Given the relatively poor ASR obtained by CWL2 attacks against stochastic ANNs, we utilised randomness in augmenting input samples in the attack procedure to create attacks which result in samples further away from the decision boundary and thus are able to mislead stochastic ANNs. The CWL2 attack solves the objective given by Equation~\ref{cweqn}. We modify this function to include an additional term that performs random augmentations on the input image, both rotations and translations, and then optimising it, gaining inspiration from \cite{eot}. Equation~\ref{modcweqn} describes our modified attack, ModCWL2.
\begin{align}
    \label{modcweqn}
    &min_\delta ||\delta||_2 + c\cdot f(x+\delta) + \frac{1}{K}\sum_{i=1}^{K} f(R(x+\delta))
\end{align}
\noindent where $K$ is the number of iterations to perform random transformations, symbolised by $R(.)$, on the input sample. Our function $R(.)$ involves first making random rotations followed by random translations. In this work, we defined the allowable range of rotation angles to $180$ degrees clockwise and counterclockwise, sampled from a uniform distribution. Also, we select at random the translation direction and pixels (integer from 0 to 10) to be applied on the image. 
This modification will induce a trade-off between resultant $L_2$ norms and ASR. 
One can understand it in the following way: performing $K$ times random transformations $R$ will turn a single sample $x$ into a cluster of $K$ samples. Moving the cluster as a whole over the decision boundary requires a larger step than moving a single sample, depending on the radius of the cluster. 




We adopt the same experimental settings as in CWL2 attacks. In total, out of 12 configurations of stochastic networks in Table~\ref{otherattackmain:otherattacks}, ModCWL2 performs in 9 configurations better compared to CWL2 in terms of ASR. For deterministic models, this attack yields also a higher ASR.


\subsection{Black-box Attacks Against Neural Networks}

\subsubsection{Boundary Attack}
The results in Table~\ref{otherattackmain:otherattacks} shows that this attack is highly effective against deterministic models, with the exception of BNN-D in PCam. However, this attack turns out to be the poorest performing for the case of stochastic ANNs. It is important to note that the boundary attack was performed on an average of queries for the same data point to counter the stochasticity.
This observation indicates that the attack method is very sensitive to stochasticity in the model.

In the case of deterministic models, the decision boundary remains stable after training due to for the same input sample. On the other hand, for stochastic ANNs, its weights and activations will vary based on a probability distribution, resulting in slightly varied predictions for the same sample at different times. 
Having a stochastic decision boundary will compromise the ability to obtain accurate feedback for the traversal of adversarial sample candidates which explains the poor performance of this attack. Section~\ref{stocboundary} illustrates this point further.





\subsubsection{Transferability Attacks}

We discuss the transferability of adversarial samples derived from ResNet18 to other architectures. This is a plausible scenario, arising when the attacker chooses a common CNN (i.e. ResNet18) as target for adversarial attacks. He or she then generates adversarial samples from the CNN, and launches them against the actual target model which is based on a different architecture. 
We chose a subset of network variants instead of the full range of models in this set of experiments as we ignored repetitive variants and also variants already highly susceptible to the standard mode of attacks.



\begin{table*}[htb]
\centering
\caption{Transferability rate of the resultant adversarial samples
generated from ResNet18 using various attack types on CIFAR-10. Only adversarial samples successful against ResNet18 were considered. 
$c=10$ for the CW-based attacks.}
\label{tab:transfer}
\resizebox{0.6\textwidth}{!}{%
\begin{tabular}{cccccc}
\hline
Attack method        & $\epsilon$ & MCSEFRON & SNN$_M$ & BSN-L  & BNN-S  \\ \hline
\multirow{3}{*}{BIM} & 8/255      & 0.1500   & 0.2400  & 0.1784 & 0.2969 \\
                     & 16/255     & 0.2900   & 0.3200  & 0.2893 & 0.4275 \\
                     & 32/255     & 0.3000   & 0.3200  & 0.3361 & 0.4126 \\\hline
CWL2                 & -          & 0.0100   & 0.0400  & 0.0822 & 0.0573 \\ \hline
ModCWL2              & -          & 0.0300   & 0.0600  & 0.0776 & 0.0759 \\ \hline
Boundary             & -          & 0.0500   & 0.1100  & 0.0647 & 0.0619 \\ \hline
\end{tabular}
}
\end{table*}

We draw the following observations based on Table~\ref{tab:transfer}.
Firstly, we observe highest transferability rates for BIM attacks. Also, as $\epsilon$ increases, so does the transferability rate. This indicates that coarser attacks are more successful in generating transferable adversarial samples, as these move into zones where misclassification is shared across models. Transferability rates are also among the highest against BNN-S. Likely this is due to the similar base architectures between BNN-S and ResNet18 as BNN-S uses ResNet18 as a structure while replacing components with binarized and stochastic counterparts.
Secondly, elaborate attacks yield low success rates against all model variants. This indicates that the decision boundaries of these explored variants are highly different compared to the ResNet18. We consider it a relevant contribution of our study, to show that stochastic model variants are moderately robust against direct transfer attacks.






\subsection{Surrogate-based Black-box Attacks}

In this section, we report the effectiveness of Surrogate-based black-box attacks, as introduced by \cite{papernot2017practical}, and propose Variance Mimicking as an improvement for the case of stochastic targets. 
In our experiments, we considered three different model variants as our target classifier (i.e. oracle), namely BSN-L, BNN-S and MCSEFRON. 
We evaluated on both the MNIST and CIFAR-10 datasets, taking 20\% of the test data to be used for training the surrogate. The remaining test data were used to evaluate the performance of the surrogate models.

As illustrated in Table~\ref{surrogateres}, the ModCWL2 attack achieves higher transferability as compared to the vanilla CWL2 attack, which is also consistent with our explanation provided in Section~\ref{augcwattack}. However, the BIM attack is in general more efficient in attaining transferable adversarial samples than the other, more complex, attack methods. This observation, together with the results in Table~\ref{tab:transfer}, suggests that simple iterative attacks are better performing for attacks involving transferability. 

\begin{table*}[htbp]
\centering
\caption{\label{surrogateres} Surrogate-based attack success rates using the \textit{ResNet18} model as surrogate and launching against the respective oracles. 100 samples were used for each experiment.}
\resizebox{0.65\textwidth}{!}{%
\begin{tabular}{lllccc}
\hline
\multirow{2}{*}{Dataset}  & \multirow{2}{*}{Attack Method} & \multicolumn{1}{c}{\multirow{2}{*}{$\epsilon$/$c$}} & \multicolumn{3}{c}{Oracle} \\
                          &                                & \multicolumn{1}{c}{}                                & MCSEFRON & BSN-L  & BNN-S  \\ \hline
\multirow{3}{*}{MNIST}    & BIM                            & $\epsilon=32/255$                                   & 0.2526   & 0.3898 & 0.4219 \\
                          & CWL2                           & $c=10$                                              & 0.0500   & 0.0474 & 0.0304 \\
                          & ModCWL2                        & $c=10$                                              & 0.2300   & 0.5831 & 0.2335 \\ \hline
\multirow{3}{*}{CIFAR-10} & BIM                            & $\epsilon=32/255$                                   & 0.7100   & 0.4337 & 0.2664 \\
                          & CWL2                           & $c=10$                                              & 0.6701   & 0.2034 & 0.1302 \\
                          & ModCWL2                        & $c=10$                                              & 0.6700   & 0.2094 & 0.1422 \\ \hline
\end{tabular}
}
\end{table*}


As an extension, we consider the ASR when a stochastic model (i.e. BSN-L) is used as a surrogate, which to the best of our knowledge, has not been explored before.
Prior art frequently discussed surrogate-based attacks with deterministic models (e.g. ResNets) as surrogates \cite{papernot2017practical,Galloway2017,cheng2019improving} instead.
We hypothesise that, using a deterministic surrogate against stochastic oracles, results in a reduced ASR due to the difficulty to account for the stochastic variance of the decision boundary.
As querying the stochasticity of a target is trivial (through multiple queries of the same sample), the attacker can easily employ a stochastic surrogate.

\begin{figure}[htb]
    \centering
    \includegraphics[trim={2.0cm 10.3cm 2.0cm 11.5cm},clip,width=\linewidth]{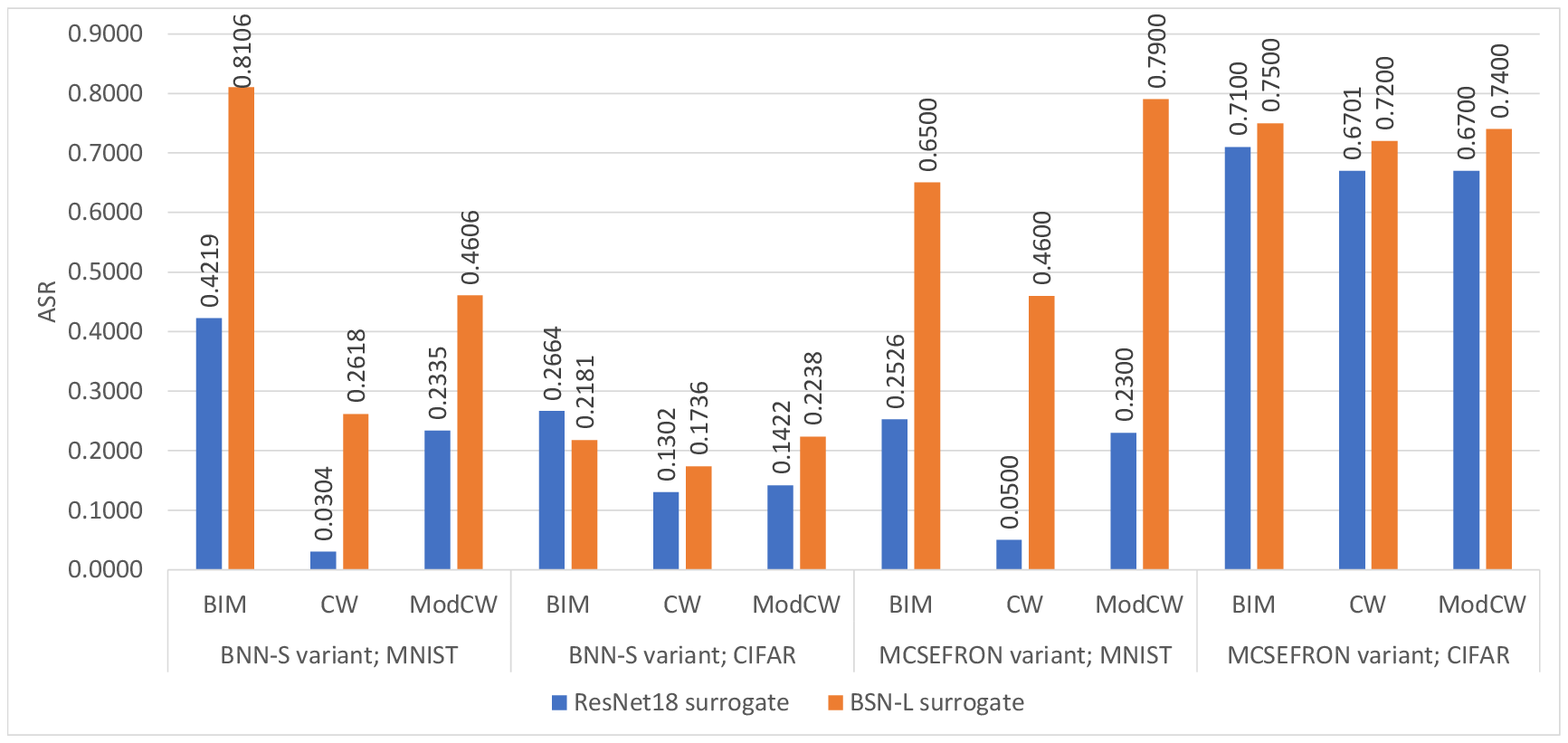}
    \caption{Attack success comparison between ResNet18 (blue) vs BSN-L (orange) models as surrogates when targeting \textit{BNN-S} and \textit{MCSEFRON}. 100 samples were used. $c=10$ for CW-based attacks and $\epsilon=32/255$ for BIM.}
    \label{surrogatevanillabsn}
\end{figure}

Making observations from Figure~\ref{surrogatevanillabsn}, employing stochastic ANNs as surrogates increases the ASR against stochastic and even deterministic targets, with the exception of BIM on CIFAR-10 using a BNN-S oracle. We postulate that using stochastic ANNs acts as a regularizer to prevent adversarial perturbations from becoming too small, similar to the concept for ModCWL2 (see Section~\ref{augcwattack}). This increases the chances of fooling ANNs whenever there is no highly accurate approximation of the decision boundaries by the surrogates. 

\begin{figure}[htb]
    \centering
    \includegraphics[trim={2.0cm 9.3cm 3.0cm 9.5cm},clip,width=\linewidth]{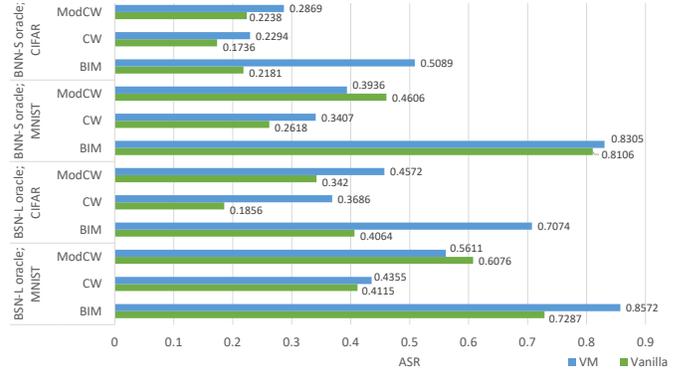}
    \caption{Attack success comparison between Vanilla (green) vs VM (blue) surrogate training procedures when using \textit{BSN-L} as surrogate. 100 samples were used. $c=10$ for CW-based attacks and $\epsilon=32/255$ for BIM.}
    \label{varmic}
\end{figure}

\subsubsection{Training Stochastic Surrogates through Variance Mimicking}

The goal of surrogate-based attacks is to approximate the decision boundary of the oracle $O$. When faced against stochastic targets, approximating the prediction variance of the oracle is equally important. This leads us to our proposed method of Variance Mimicking (VM).
Assuming the attacker uses a stochastic surrogate, he or she could inject noise in the surrogate training dataset before querying the oracle for hard labels (annotated by $\overline{O}(.)$).
We assume that in the black-box setting, we can obtain only hard labels for a predicted class, but not the underlying logits. In this case, a small perturbation $\delta$ might be insufficient to induce label flips in the prediction (i.e. $\overline{O}(x)=\overline{O}(x+\delta)$ for some input $x$), thereby defeating the point of measuring variance. To mitigate that, the attacker calculates the variance $\sigma_{B}^2$ within a mini-batch of data $B$, specific to the channel, width and height dimensions (each dimension has its own variance), and perturbs $B$ with $\mathcal{N}(0,\,\sigma_{B}^{2})$ for $m$ separate times, yielding $m*B$ samples. The attacker then trains the surrogate on this expanded mini-batch with their respective labels provided by the oracle, as usual like in \cite{papernot2017practical}. 


As shown in Figure~\ref{varmic}, the VM procedure outperforms the naive approach, in terms of ASR, exception for two cases. While these results do not outperform the white-box attacks (see Table~\ref{otherattackmain:otherattacks}), given the difficulty of black-box attacks in above realistic setting, we consider it a positive result showing the moderate susceptibility of stochastic ANNs and the success of our proposed method.

\section{Adversarial Hardening Through Stochastic Architecture Mixtures}
\label{stocswitch}

In the previous sections, we observed that several network architectures (i.e. stochastic ANNs) appear to be moderately robust against transferability attacks.
Inspired by this, a defender could employ stochastic switching of a mixture of neural networks with differing architectures to circumvent adversarial attack attempts. At inference time, the defender chooses at random a neural network to be used to evaluate the input sample. This is a special case of drawing a distribution over networks from e.g.~a Dirichlet prior. We explore three different selected combinations of ensembles, 1) ResNet18 with BSN-L, 2) ResNet18 with BNN-S, and 3) ResNet18 with BSN-L and BNN-S. Here, we investigate the ASR in attacking against such ensembles. In our experiments, we applied the BIM attack due to its good performance against stochastic networks, using the mean of the gradients with respect to the input across the ensemble of models. This is inspired by \cite{He2017}. While they considered ensembles of conventional ANNs, we explore a stochastic mixture of differing architectures. 

\begin{figure}[htb]
    \centering
    \includegraphics[trim={2.4cm 9.7cm 2.4cm 11.2cm},clip, width=0.9\linewidth]{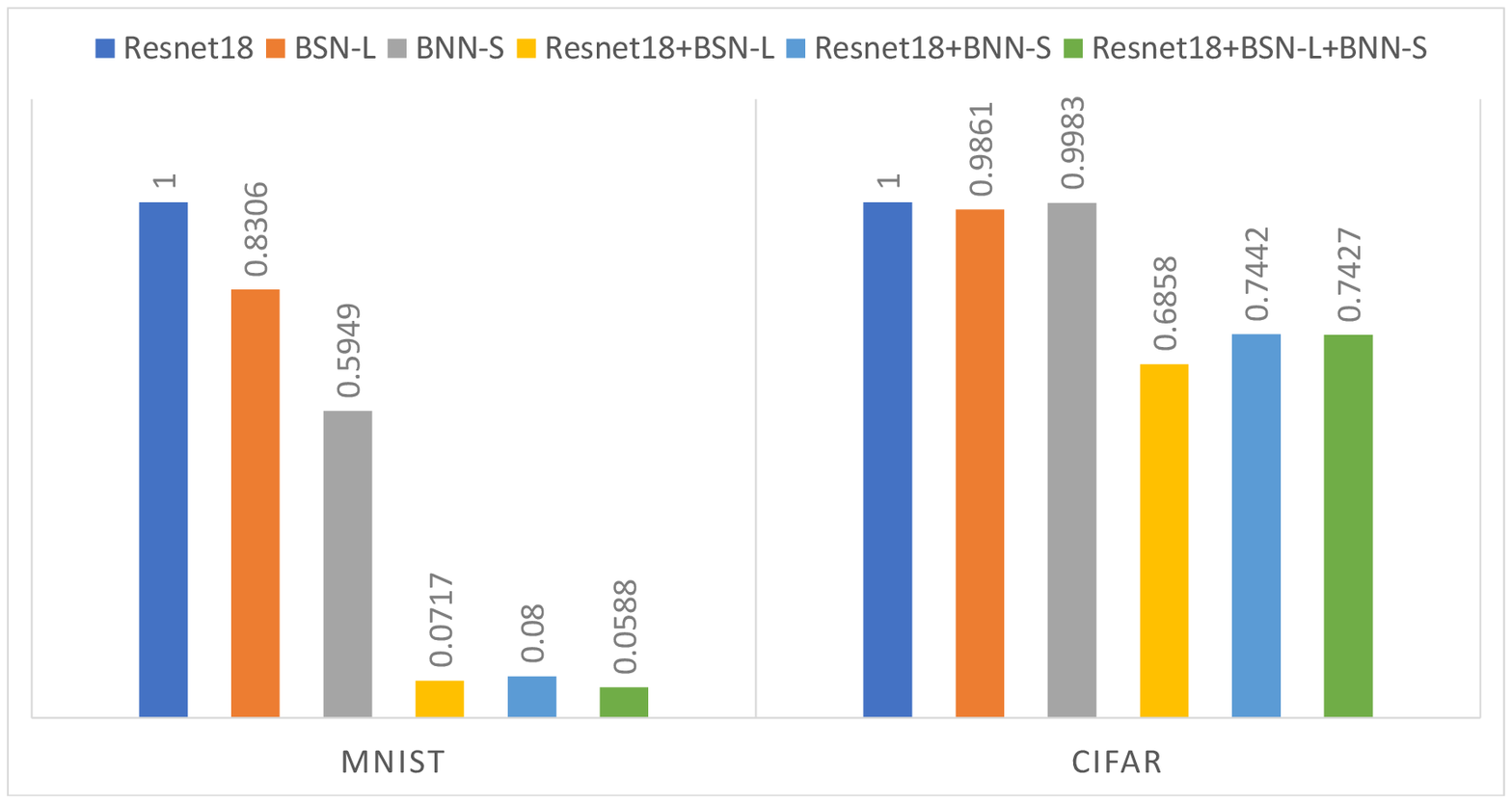}
    \caption{ASR (in [0,1]) against single, un-switched ANNs (left three bars for each data set; \textit{Group A}) and stochastically switched ANNs (ResNet18+BSN-L, ResNet18+BNN-S, ResNet18+BSN-L+BNN-S, right three bars for each data set; \textit{Group B}) for the BIM attacks at $\epsilon=32/255$, for MNIST and CIFAR-10. 
    Results for single ANNs taken from Table~\ref{otherattackmain:otherattacks} under BIM attack.}
    \label{xarchiasr}
\end{figure}

As evident in Figure~\ref{xarchiasr}, there is a decrease in ASR when stochastic switching mechanisms were implemented. ASR reported for Group B were consistently lower than their Group A counterparts. This suggests a plausible hardening mechanism that a defender can put in place to improve the resilience against adversarial attacks. This is an interesting finding of our work which warrants further investigation with more complex datasets and a larger ensemble.

\section{Discussion}
\label{discuss}

\subsection{Susceptibility of Stochastic ANNs Against Adversarial Attacks}
\label{stocsuscep}
In white-box settings, stochastic networks are almost equally very vulnerable as conventional ANNs, when BIM is used with sufficient strength. It is the simplest of all considered attacks. Its advantage for stochastic networks is that it does not attempt to stay close to the decision boundary as explicitly enforced in boundary attacks and CWL2 attacks. For stochastic networks the decision boundary is defined only in an expected sense. Staying close to expected decision boundary results in a higher failure rate of adversarials. 
When attacking stochastic ANNs under realistic attack scenarios, the ASR for black-box attacks are even more pessimistic. In our experiments, we show that surrogate-based black-box attacks are the most successful as compared to the Boundary and transferability attacks. However, even the surrogate-based attack is nowhere near as successful as the weakest white-box attack in our study, CWL2.


\subsection{Understanding the impact of stochasticity}
\label{stocboundary}
\begin{figure*}[htb]
    \centering
    \begin{subfigure}[t]{0.49\textwidth}
        \centering
        \includegraphics[trim={3.6cm 1.9cm 2.3cm 2.1cm},clip, width=0.6\linewidth]{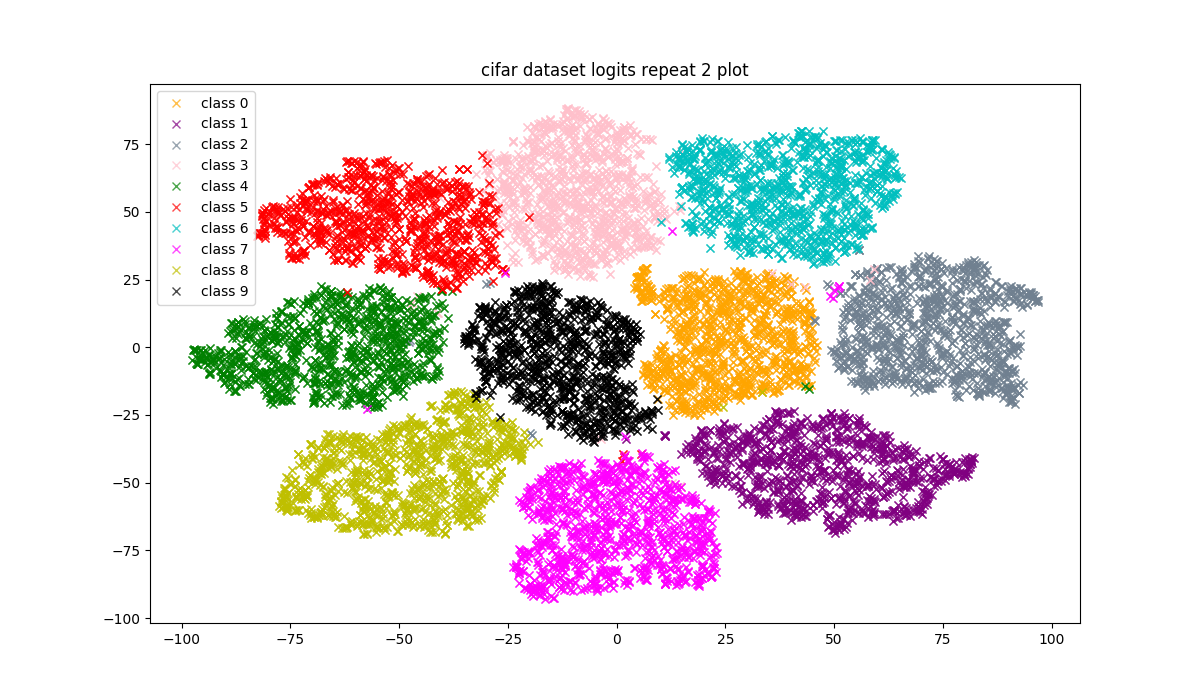}
        \caption{t-SNE of logits: ResNet18 classifier}
    \end{subfigure}
    ~ 
    \begin{subfigure}[t]{0.49\textwidth}
        \centering
        \includegraphics[trim={3.6cm 1.9cm 2.3cm 2.1cm},clip,width=0.6\linewidth]{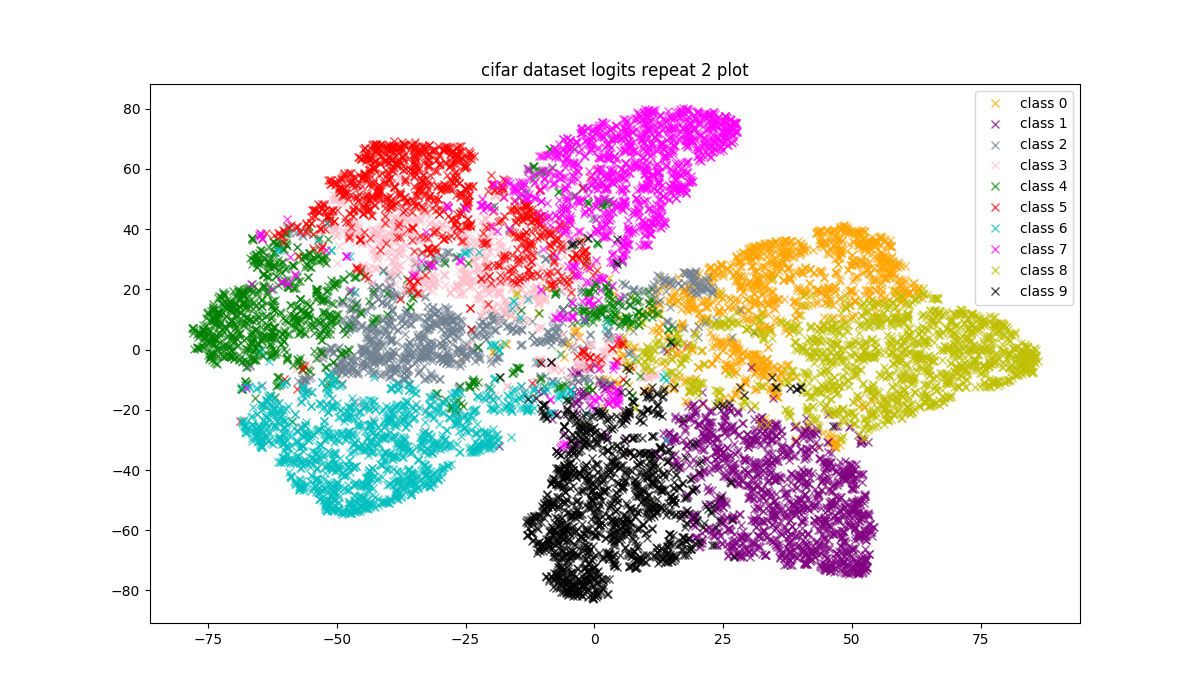}
        \caption{t-SNE of logits: BSN-L classifier}
    \end{subfigure}\\
    
        \begin{subfigure}[t]{0.48\textwidth}
        \centering
        \includegraphics[trim={3.6cm 1.9cm 2.3cm 2.1cm},clip, width=0.5\linewidth]{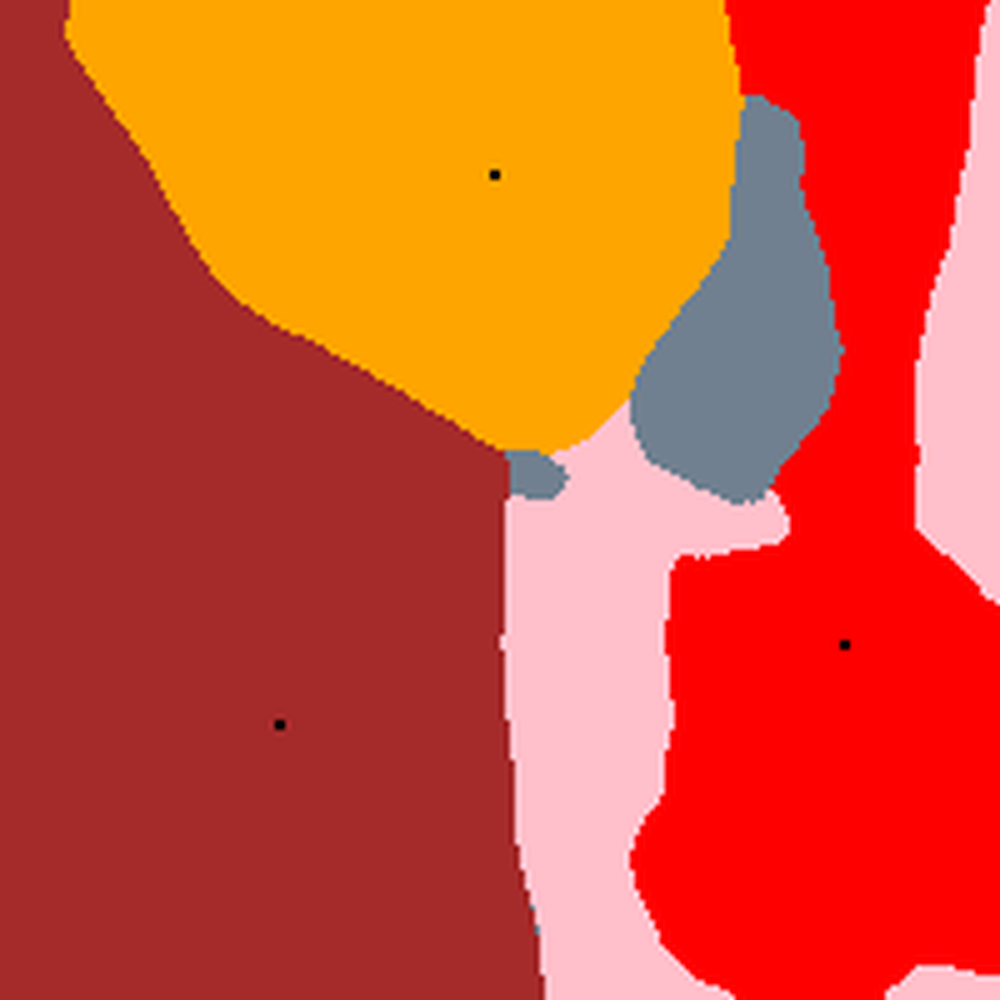}
        \caption{Predicted classes for an exemplary region: ResNet18 classifier. Prediction in each point is deterministic. No variance plotted.}
    \end{subfigure}
    ~ 
    \begin{subfigure}[t]{0.48\textwidth}
        \centering
        \includegraphics[trim={3.6cm 1.9cm 2.3cm 2.1cm},clip,width=0.5\linewidth]{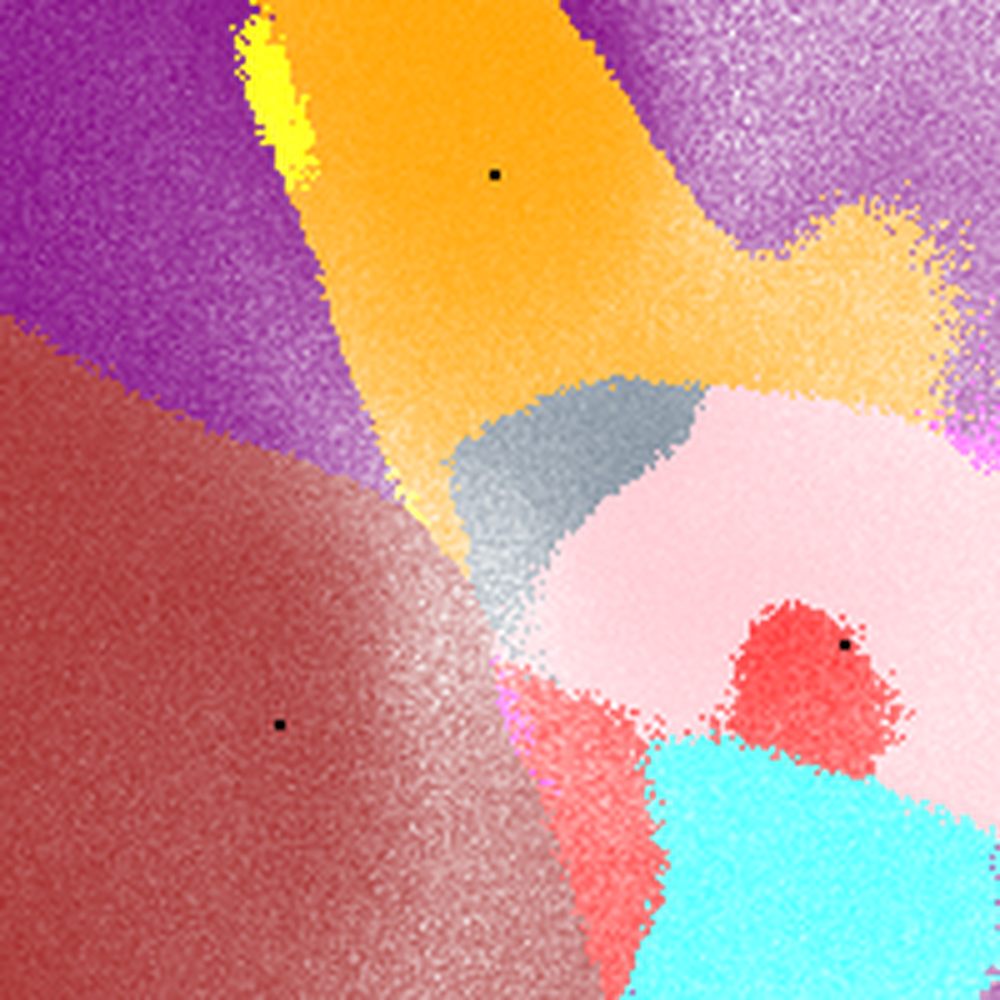}
        \caption{Predicted classes and variances for an exemplary region: BSN-L classifier. High variance appears as low gamma, resulting in fading colours.}
    \end{subfigure}
    \caption{\label{tsnevisual}Upper row: t-SNE plots of model logits from training data of CIFAR-10. Lower row: predicted classes and variances around three data points (black dots).}
\end{figure*}

We believe that the difficulty of attacking stochastic ANNs in general is due to the variance in the definition of the decision boundary.
In an attempt to explain this phenomenon, we performed t-SNE \cite{maaten2008visualizing} plots of the logits from the training samples of MNIST to highlight the difference in the decision boundary between deterministic networks (i.e. ResNet18) and stochastic networks (i.e. BSN-L). The upper row of Figure~\ref{tsnevisual} illustrates the differences of the decision boundaries between deterministic (i.e. ResNet18) and stochastic (i.e. BSN-L) models. It is clear that there is more overlap between samples of differing classes and the lack of a clear boundary in the stochastic case. 

The lower row of Figure~\ref{tsnevisual} shows for the BSN-L classifier the predicted labels and the variances, and for the Resnet18 only the predicted labels. This was computed on a section defined by a two-dimensional plane spanned by three exemplary data points. One can see two peculiarities: Firstly, the decision boundaries for BSN-L are much more spiky and noisy, as can be seen for the red labels in the lower right corner, and the purple in the upper right corner, with zones of high variance in white around them. Traversing along a decision boundary is more challenging for stochastic networks. This shows the difficulty for boundary attacks against stochastic models. Secondly, decision boundaries around the same three samples are substantially different for the two networks. The zones of high variance provide additional obstacles to correctly training surrogates. This explains the difficulties for transferability and surrogate attacks, and the reason why stochastic mixtures are able to provide the observed hardening. 

\section{Conclusion}
\label{conclude}

We investigated the adversarial robustness of a wide variety of alternative variants of neural networks(e.g. SNNs, BSNs, BNNs), across different datasets namely MNIST, CIFAR-10 and PCam in the raw input image space. In white-box settings, stochastic ANNs are vulnerable to the simple BIM and moderately robust against more elaborate adversarial attacks, different from conventional ANNs. 
For black-box settings, we observe that stochastic neural networks are robust to boundary attacks and found that a direct transfer strategy would be highly ineffective, for elaborate attacks. However, surrogate-based transfer attacks show promise in overcoming ANNs, especially when a stochastic surrogate is used. Furthermore, we proposed variance mimicking as improved surrogate training, which specially targets stochastic ANNs by using stochastic surrogates and show that it achieves the highest ASR among all our black-box attacks explored in this work. Finally, we show that an ensemble utilising a stochastic switch of networks for inference be employed for the goal of hardening neural networks against adversarial attacks, and found that the ASR decreases. While the change in MNIST is more pronounced than that of CIFAR-10, stochasticity is an option to be considered in further research.

\section{Acknowledgements}

This work was supported by both ST Electronics and the National Research Foundation (NRF), Prime Minister’s Office, Singapore under Corporate Laboratory @ University Scheme (Programme Title: STEE Infosec-SUTD Corporate Laboratory). Alexander Binder also gratefully acknowledges the support by PIE-SGP-AI-2018-01.

\bibliographystyle{IEEEtran}
\bibliography{ms.bib}

\appendix

\subsection{Details on Image Classifier Training}

\subsubsection{ResNet18}
\label{resnet18}
We trained the MNIST model for 20 epochs from scratch while loading pre-trained versions of ResNet18 (pretrained on ImageNet) and fine-tuning them for CIFAR-10 and PCam. We selected the best performing set of weights based on the test dataset.
For the case of MNIST, where the images are single channelled, we had to replace the initial convolution layer with one that has a input channel of 1 instead of 3, since the original model on torchvision was trained on ImageNet.

During training, we performed data augmentation on the training set and also normalised the data to between 0 and 1. For the case of CIFAR-10 and PCam, we then performed colour normalisation by subtracting the mean and scaling to unit variance using the mean and standard deviation calculated from a subset of 50000 training samples from each dataset respectively. We did not perform any colour normalisation for MNIST.

\begin{table}[htb]
\centering
\caption{Hyperparameters used for training the ResNet18 model.}
\label{hyperparams}
\begin{tabular}{cccc}
\hline
                  & MNIST   & CIFAR-10 & PCam \\ \hline
Optimiser          & SGD   & Adam \cite{kingma2014adam} & Adam \\ \hline                   
Learning Rate (LR) & 0.001   & 0.001 & 0.001 \\ \hline
Weight decay (WD)      & 0.00001 & 0.0001 & 0.0001 \\ \hline
LR decay step (Step)     & 5       & 3  &  3  \\ \hline
LR decay factor ($\gamma$)    & 0.5     & 0.95  & 0.95  \\ \hline
Batch size (BS)        & 128     & 256  & 256  \\ \hline
\end{tabular}
\end{table}

\subsubsection{Mozafari et al. ($SNN_m$)}

The training code for $SNN_m$ that we used was adopted from the SpykeTorch \cite{spyketorch} library, following the tutorial provided. Each convolution layer was trained independently and chronologically via the STDP learning rule. For the last convolution layer, Reward-STDP was used instead where correct decisions were rewarded while incorrect decisions were punished. More details about the training procedure can be found in \cite{Mozafari2019}. Table \ref{mozafariparams} describes the hyperparameters we used and other hyperparameters that were not mentioned here indicates that the default values were used.

\begin{table*}[htb]
\centering
\caption{Hyperparameters used for training $SNN_m$. Threshold of $\infty$ indicates that the maximum value was taken.}
\label{mozafariparams}
\begin{tabular}{ccccc}
\hline
Dataset                  & Layer    & No. of feature maps & Kernel window & Threshold           \\ \hline
\multirow{3}{*}{MNIST}   & 1st conv & 30                  & (5,5,6)       & 15                  \\
                         & 2nd conv & 250                 & (3,3,30)      & 10                  \\
                         & 3rd conv & 200                 & (5,5,250)     & $\infty$ \\ \hline
\multirow{3}{*}{CIFAR-10} & 1st conv & 300                 & (5,5,6)       & 15                  \\
                         & 2nd conv & 800                 & (3,3,300)     & 10                  \\
                         & 3rd conv & 500                 & (5,5,800)     & $\infty$ \\ \hline

\end{tabular}
\end{table*}

\subsubsection{Binary Stochastic Neuron (BSN) Models}
We performed the similar preprocessing steps as described for the case of ResNet18. For our BSN models, we used the Straight-Through (ST) estimator in the backpropagation of the gradients. Slope annealing was done in conjunction with the ST estimator for the BSN-2 and BSN-4 variants only. More details about the slope annealing trick can be found in \cite{chung2016hierarchical}.
Table \ref{bsnparams} summarises some of the hyperparameters that we used in our experiments for the BSNs. As mentioned in the main text, we used a batch size of 128 across the settings and used Adam optimiser for the BSN-4 and BSN-L variants while using SGD for BSN-2. 
\begin{table*}[htb]
\centering
\caption{Hyperparameters used for training the BSNs.}
\label{bsnparams}
\begin{tabular}{cccccccccc}
\hline
         & \multicolumn{3}{c}{BSN-2} & \multicolumn{3}{c}{BSN-4} & \multicolumn{3}{c}{BSN-L} \\ \hline
         & MNIST  & CIFAR-10 & PCam  & MNIST  & CIFAR-10 & PCam  & MNIST  & CIFAR-10 & PCam  \\ \hline
LR       & 0.01   & 0.01    & 0.01   & 0.0001 & 0.0001  & 0.0001 & 0.001  & 0.01    & 0.01   \\ \hline
WD       & 0.0001 & 0.0001   & 0.0001 & 1E-6   & 1E-6    & 1E-6   & 1E-5   & 0.0001  & 0.0001 \\ \hline
Step     & 5      & 5       & 5      & 5      & 5       & 5      & 50      & 50      & 5      \\ \hline
$\gamma$ & 0.75   & 0.75    & 0.75   & 0.75   & 0.75    & 0.75   & 0.1   & 0.5     & 0.75   \\ \hline
\end{tabular}
\end{table*}

Recall that BSN-2 indicates a simple 2-layered Multi-layer Perceptron (MLP) while BSN-4 indicates a 4-layered MLP variant. The difference between the standard MLP and our BSN here is that all hidden activations are modified to perform binarization of activations in a stochastic manner.

\begin{table}[htb]
\centering
\caption{Dimensions of the Fully Connected (FC) layers for the BSN-2 model across the datasets.}
\label{bsn2modelarchi}
\begin{tabular}{llll}
\hline
             & MNIST & CIFAR-10 & PCam \\ \hline
Input layer  & 784   & 3072    & 3072  \\ \hline
Hidden layer & 100   & 100     & 100   \\ \hline
Output layer & 10    & 10      & 2     \\ \hline
\end{tabular}
\end{table}

\begin{table}[htb]
\centering
\caption{Dimensions of the Fully Connected (FC) layers for the BSN-4 model across the datasets.}
\label{bsn4modelarchi}
\begin{tabular}{llll}
\hline
               & MNIST & CIFAR-10 & PCam \\ \hline
Input layer    & 784   & 3072    & 3072  \\ \hline
Hidden layer 1 & 600   & 2000    & 2000  \\ \hline
Hidden layer 2 & 400   & 750     & 750   \\ \hline
Hidden layer 3 & 200   & 100     & 100   \\ \hline
Output layer   & 10    & 10      & 2     \\ \hline
\end{tabular}
\end{table}

\subsubsection{Training the Classifiers}
\label{trainingclfs}

For the case of MCSEFRON, we used five receptive fields (RFs) and a learning rate of 0.1 for MNIST while using three RFs and a learning rate of 0.5 for CIFAR-10. The other hyperparameters were set at their default values. We used the authors' implementation of MCSEFRON in Python\footnote{https://github.com/nagadarshan-n/MC-SEFRON} for training. 
In training MCSEFRON, we performed sub-sampling strategies on the training data. We used the first batch of training data of CIFAR-10; we used the first 30000 samples of PCam.

As mentioned in Section~2.3 of our main text, for the case of SNN$_M$, the model's input images are
preprocessed by the DoG filters. The number of DoG filters used will determine the input channel of the first convolution layer in SNN$_M$. Hence, for a three-channelled image (e.g. CIFAR-10), we first take the mean of the channels to convert the images to a single channel, prior to passing them to the DoG filters. Unfortunately for this model, we could not find a suitable set of hyperparameters that performs reasonably on the PCam dataset. While training, we noticed that the outputs of the network was consistently the same, regardless of the number of training iterations. Hence, we could not report the Adversarial Success Rates (ASRs) and their respective norms for the attacks against SNN$_M$ using the PCam dataset.

For BSNs, we used a batch size of 128 and used Adam optimizer for the BSN-4 and BSN-L variants while using Stochastic Gradient Descent (SGD) for BSN-2. 
The other hyperparameters we used can be found in Table~\ref{bsnparams}.
We adapted the code from this GitHub repository\footnote{https://github.com/Wizaron/binary-stochastic-neurons}, with the network definition of the BSN-L architecture in PyTorch requiring modification on all intermediate activations with BSN modules. 

For the BNNs, we used the same hyperparameters across the various datasets and models and adapted the code from this GitHub repository\footnote{https://github.com/itayhubara/BinaryNet.pytorch}, which was originally used by the authors in \cite{Hubara2016}. We used a learning rate of 0.005 and weight decay of 0.0001 with a batch size of 256. We also used the Adam optimiser to train our models for 20 epochs in MNIST, 150 epochs in CIFAR-10 and 50 epochs in PCam. We manually set the learning rate to 0.001 at epoch 101 and 0.0005 at epoch 142, following the authors in \cite{Hubara2016}. For BNN-D and BNN-S, we used the ResNet18 architecture as the structure of the network, while the binarization of the weights and activations will only occur at the forward pass.

\subsection{Evaluating Stochastic ANNs against Adversarial Attacks}
\label{evalstoc}
As stochastic ANNs are parameterised by a probability distribution, evaluating against such model variants requires multiple runs of the same instance for more reliable experiment results.
For all experiments performed on stochastic networks, we obtain the ASR by launching the same adversarial sample 100 times and collecting the mean ASR per sample. We then took the average of this mean as our final average across our sub-samples as our final reported value. 

\subsection{Modifying SNN implementation for Adversarial Attack Evaluation}
\label{modsnn}
As SNNs are inherently very different from conventional ANNs, there is a need to adapt the original implementation of the SNNs to fit our purposes. We made two modifications in our work.
First, because there might be instances in which non-differentiable operations were performed (i.e. sign function), when adapting such SNNs for our use, we replaced the built-in sign functions with our custom sign function, which performs the same operation but allows gradients to pass through in a straight through fashion in the backward pass. This ensures that the gradients are non-zeros everywhere. Also, since we examined SNNs that were trained via STDP, such a change does not violate the learning rule of the SNNs. Furthermore, as we are only interested in the behaviour of such models when faced with adversarial samples, we extracted the critical parts of the network (i.e. decision-making forward pass) only in our adaptation.

Secondly, as SNNs make decisions based on either earliest spike times or maximum internal potentials, their outputs are more commonly a single valued integer, depicting the predicted class. However, for attacks to be done on such networks, we require logits of networks. Hence, we simulated logits in our modification by using the post-synaptic spike times for the case of MCSEFRON and the potentials for the case of SNN$_M$ for all of the classes. When spike times were used, we took the negative of spike times so that the max of the vector of spike times correspond to the actual prediction.
We note that we make the aforementioned modifications under the scenario of an adaptive attacker perspective to form a more well-rounded analysis.

\subsection{Full Experiment Results for BIM Attacks}
\label{fullbim}
Tables~\ref{bimattack} and \ref{bimattackl2norms} shows the experiment results we obtained when we vary the attack strength for the BIM attack. We used three different values of $\epsilon$, namely $8/255$, $16/255$ and $32/255$. It is clear that BNN-D is highly susceptible to even the simple adversarial attacks with low attack strength, as the ASR for BNN-D is higher than that of ResNet18 while having lower $L_2$ norms per pixel at $\epsilon=8/255$.

\begin{table*}[htb]
\centering
\caption{Adversarial success rate and mean $L_2$ norms per pixel for the BIM attack. $\epsilon$ is the attack strength.}
\begin{subtable}[h]{\textwidth}
\centering
\caption{ASR (in {[}0,1{]}) of the different variants of models. Step size of 0.05 and 100 iterations were performed. 500 samples were sampled for each experiment.}
\label{bimattack}
\resizebox{0.9\textwidth}{!}{%
\begin{tabular}{clcccccccc}
\hline
$\epsilon$ & Dataset & Resnet18 & $SNN_m$ & MCSEFRON & BSN-2  & BSN-4  & BSN-L  & BNN-D & BNN-S  \\ \hline
\multirow{3}{*}{8/255}      & MNIST   & 0.1900    & 0.0260   & 0.0540    & 0.0981 & 0.1782 & 0.0299 & 0.9920 & 0.2466 \\
                            & CIFAR-10 & 0.9900   & 0.3580   & 0.9800    & 0.9516 & 0.9891 & 0.9273 & 1.0000 & 0.9891 \\
                            & PCam    & 0.9440    & -       & 0.2440    & 0.9063 & 0.8129 & 0.6224 & 0.8590 & 0.9599 \\ \hline
\multirow{3}{*}{16/255}     & MNIST   & 0.7960    & 0.0560   & 0.1560    & 0.3663 & 0.6520 & 0.1998 & 1.0000 & 0.4288 \\
                            & CIFAR-10 & 1.0000    & 0.5340   & 0.9980    & 0.9900 & 0.9980 & 0.9900 & 1.0000 & 0.9981 \\
                            & PCam    & 0.9740    & -       & 0.4160    & 0.9856 & 0.9349 & 0.8111 & 0.9260 & 0.9751 \\ \hline
\multirow{3}{*}{32/255}     & MNIST   & 1.0000    & 0.1200   & 0.2940    & 0.8741 & 0.9556 & 0.8306 & 1.0000 & 0.5949 \\
                            & CIFAR-10 & 1.0000    & 0.6940   & 0.9980    & 0.9874 & 0.9880 & 0.9861 & 1.0000 & 0.9983 \\
                            & PCam    & 1.0000   & -       & 0.5340    & 0.9870 & 0.9549 & 0.9154 & 0.9740 & 0.9739 \\ \hline
\end{tabular}
}
\end{subtable}
\hfill
\begin{subtable}[h]{\textwidth}
\centering
\caption{Mean $L_2$ norms per pixel between the original image and its perturbed adversarial image of the different variants of models. Note that the values reported have been scaled up by a factor of 1000.
}
\label{bimattackl2norms}
\resizebox{0.9\textwidth}{!}{%
\begin{tabular}{clcccccccc}
\hline
$\epsilon$ & Dataset & Resnet18 & $SNN_m$ & MCSEFRON & BSN-2  & BSN-4  & BSN-L  & BNN-D  & BNN-S  \\ \hline
\multirow{3}{*}{8/255}      & MNIST   & 0.8301   & 0.8201  & 0.4681   & 0.8510 & 0.8815 & 0.8387 & 0.8289 & 0.8223 \\
                            & CIFAR-10 & 0.5524   & 0.5338  & 0.5581   & 0.5298 & 2.1455 & 0.5038 & 0.5483 & 0.5445 \\
                            & PCam    & 0.5530   & -       & 0.5324   & 0.5511 & 0.5584 & 0.5339 & 0.5461 & 0.5540 \\ \hline
\multirow{3}{*}{16/255}     & MNIST   & 1.4925   & 1.4212  & 0.9072   & 1.5920 & 2.1455 & 1.5886 & 1.3604 & 1.4794 \\
                            & CIFAR-10 & 0.8809   & 0.9129  & 0.8866   & 0.8820 & 0.8882 & 0.8196 & 0.8767 & 0.8664 \\
                            & PCam    & 0.8452   & -       & 0.8600   & 0.8767 & 0.8763 & 0.8671 & 0.8698 & 0.8866 \\ \hline
\multirow{3}{*}{32/255}     & MNIST   & 2.1667   & 2.4142  & 1.4126   & 2.5725 & 2.2894 & 2.6324 & 1.5606 & 2.2744 \\
                            & CIFAR-10 & 0.9318   & 1.3968  & 0.8924   & 1.1485 & 1.0247 & 1.0947 & 0.9606 & 1.0407 \\
                            & PCam    & 0.9794   & -       & 1.4248   & 0.9491 & 0.9426 & 1.0308 & 1.0343 & 0.9882 \\ \hline
\end{tabular}
}
\end{subtable}
\end{table*}

\subsection{Full Experiment Results for CW-based Attacks Against Stochastic ANNs}

We present our full experiment results against stochastic ANNs at varying $c$ values, defined in both Equations~2 and 3 (see main text), at $c=0.01, 0.1, 1, 10$ while keeping the other hyperparameters constant.

\begin{table*}[htb]
\centering
\caption{ASR and mean $L_2$ norms per pixel for CW-based attacks on the stochastic ANNs. Average inference policy was used.}
\label{tab1:main}
\begin{subtable}[h]{\textwidth}
\centering
\caption{ASR (in [0,1]) across the different datasets for stochastic ANNs.}
\label{tab1:cwattackfullstoc}
\resizebox{0.91\textwidth}{!}{%
\begin{tabular}{cccccccccc}
\hline
$c$                                     & Dataset                     & \multicolumn{2}{c}{BSN-2}                               & \multicolumn{2}{c}{BSN-4}                               & \multicolumn{2}{c}{BSN-L}                               & \multicolumn{2}{c}{BNN-S}                               \\
                                        &                             & CWL2                       & ModCWL2                    & CWL2                       & ModCWL2                    & CWL2                       & ModCWL2                    & CWL2                       & ModCWL2                    \\ \hline
\multirow{3}{*}{0.01}                   & MNIST                       & 0.0016                     & 0.8551                     & 0.0000                     & 0.8238                     & 0.0011                     & 0.6751                     & 0.2380                     & 0.7356                     \\
                                        & CIFAR-10                     & 0.1231                     & 0.9462                     & 0.2162                     & 0.6637                     & 0.1879                     & 0.8034                     & 0.6226                     & 0.7894                     \\
                                        & PCam                        & 0.0306                     & 0.4275                     & 0.0157                     & 0.3467                     & 0.0241                     & 0.2428                     & 0.2888                     & 0.4706                     \\ \hline
\multirow{3}{*}{0.1}                    & MNIST                       & 0.0199                     & 0.7852                     & 0.0365                     & 0.6781                     & 0.0021                     & 0.6075                     & 0.1207                     & 0.5194                     \\
                                        & CIFAR-10                     & 0.2335                     & 0.9163                     & 0.3505                     & 0.6054                     & 0.3064                     & 0.6251                     & 0.4222                     & 0.6731                     \\
                                        & PCam                        & 0.0585                     & 0.3297                     & 0.0473                     & 0.3228                     & 0.0303                     & 0.2616                     & 0.1265                     & 0.4157                     \\ \hline
\multirow{3}{*}{1}                      & MNIST                       & 0.3435                     & 0.4718                     & 0.5569                     & 0.6860                     & 0.3463                     & 0.5889                     & 0.0875                     & 0.4142                     \\
                                        & CIFAR-10                     & 0.4533                     & 0.9219                     & 0.6272                     & 0.7125                     & 0.6079                     & 0.7268                     & 0.4405                     & 0.4875                     \\
                                        & PCam                        & 0.2233                     & 0.3878                     & 0.1547                     & 0.2799                     & 0.1550                     & 0.2117                     & 0.0888                     & 0.2246                     \\ \hline
\multirow{3}{*}{10}  & MNIST                & 0.8126               & 0.8075               & 0.8487               & 0.8655               & 0.7772               & 0.7853               & 0.2998               & 0.2934               \\
                     & CIFAR-10              & 0.5820               & 0.9402               & 0.7973               & 0.8545               & 0.7892               & 0.8509               & 0.6603               & 0.6926               \\
                     & PCam                 & 0.4932               & 0.6024               & 0.4755               & 0.5471               & 0.4772               & 0.4627               & 0.3216               & 0.3956               \\ \hline
\end{tabular}
}
\end{subtable}

\begin{subtable}[h]{\textwidth}
\centering
\caption{Mean $L_2$ norms per pixel between the original samples and their adversarial counterparts, across the different datasets for stochastic ANNs. Note that the metrics reported here are scaled up by a factor of 1000.}
\label{tab1:cwattackl2normsfullstoc}
\resizebox{0.91\textwidth}{!}{%
\begin{tabular}{cccccccccc}
\hline
$c$                   & Dataset & \multicolumn{2}{c}{BSN-2} & \multicolumn{2}{c}{BSN-4} & \multicolumn{2}{c}{BSN-L} & \multicolumn{2}{c}{BNN-S} \\
                      &         & CWL2        & ModCWL2     & CWL2        & ModCWL2     & CWL2        & ModCWL2     & CWL2        & ModCWL2     \\ \hline
\multirow{3}{*}{0.01} & MNIST   & 0.0271      & 8.6013      & -           & 8.7636      & 0.0221      & 9.3496      & 0.0021      & 9.4627      \\
                      & CIFAR-10 & 0.0028      & 22.4262     & 0.0048      & 1.1484      & 0.0029      & 2.3630      & 0.0022      & 2.0826      \\
                      & PCam    & 0.0044      & 2.1047      & 0.0048      & 1.9568      & 0.0035      & 2.8278      & 0.0077      & 2.8576      \\ \hline
\multirow{3}{*}{0.1}  & MNIST   & 0.2427      & 8.0369      & 0.5680      & 7.0468      & 0.3116      & 9.3235      & 0.0208      & 8.3933      \\
                      & CIFAR-10 & 0.0317      & 21.2624     & 0.0449      & 0.5288      & 0.0365      & 1.2123      & 0.0214      & 1.4288      \\
                      & PCam    & 0.0407      & 1.0505      & 0.0415      & 1.6038      & 0.0304      & 2.3510      & 0.0214      & 2.2465      \\ \hline
\multirow{3}{*}{1}    & MNIST   & 1.2384      & 2.4610      & 1.1406      & 2.3033      & 1.3168      & 4.1437      & 0.2126      & 7.0133      \\
                      & CIFAR-10 & 0.0571      & 8.0159      & 0.0743      & 0.2318      & 0.0769      & 0.3323      & 0.0748      & 0.2212      \\
                      & PCam    & 0.1079      & 0.2974      & 0.1562      & 0.3123      & 0.1785      & 0.3861      & 0.0980      & 0.4213      \\ \hline
\multirow{3}{*}{10}   & MNIST   & 2.1460      & 2.2873      & 2.4674      & 2.4644      & 2.5942      & 2.6049      & 1.4761      & 1.7559      \\
                      & CIFAR-10 & 0.0862      & 5.9525      & 0.0867      & 0.1908      & 0.0916      & 0.2053      & 0.0878      & 0.1048      \\
                      & PCam    & 0.5155      & 1.0358      & 0.8428      & 0.8890      & 1.0637      & 1.0759      & 0.5285      & 0.5325      \\ \hline
\end{tabular}
}
\end{subtable}
\end{table*}


\begin{figure*}[htb]
    \centering
    \begin{subfigure}[t]{0.45\textwidth}
        \centering
        \includegraphics[trim={2cm 10.cm 2.0cm 12.2cm},clip, width=1\linewidth]{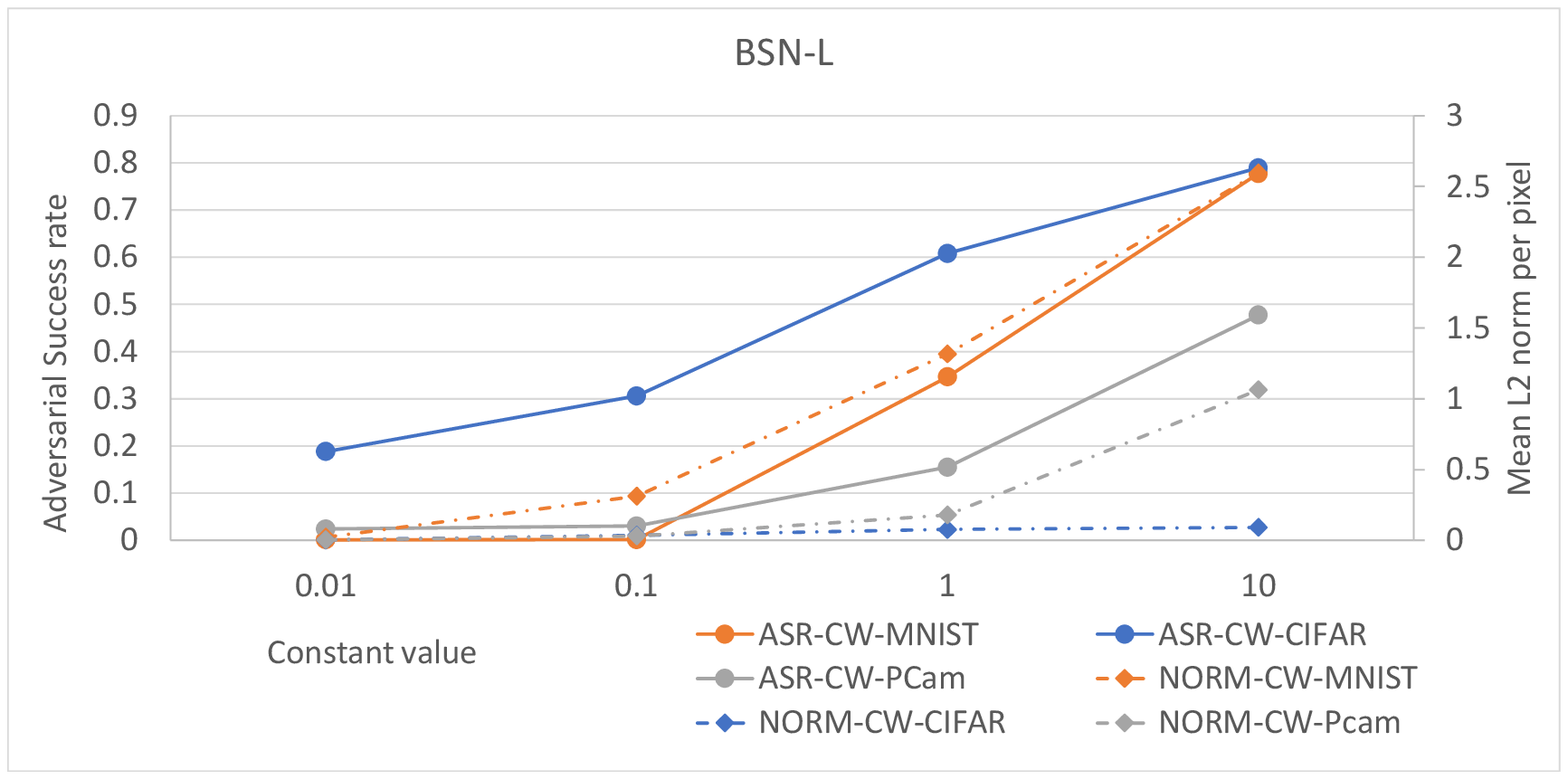}
        \caption{CWL2 attack}
        \label{cvaluetrenda}
    \end{subfigure}%
    ~ 
    \begin{subfigure}[t]{0.45\textwidth}
        \centering
        \includegraphics[trim={2cm 10.cm 2.0cm 12.2cm},clip,width=1\linewidth]{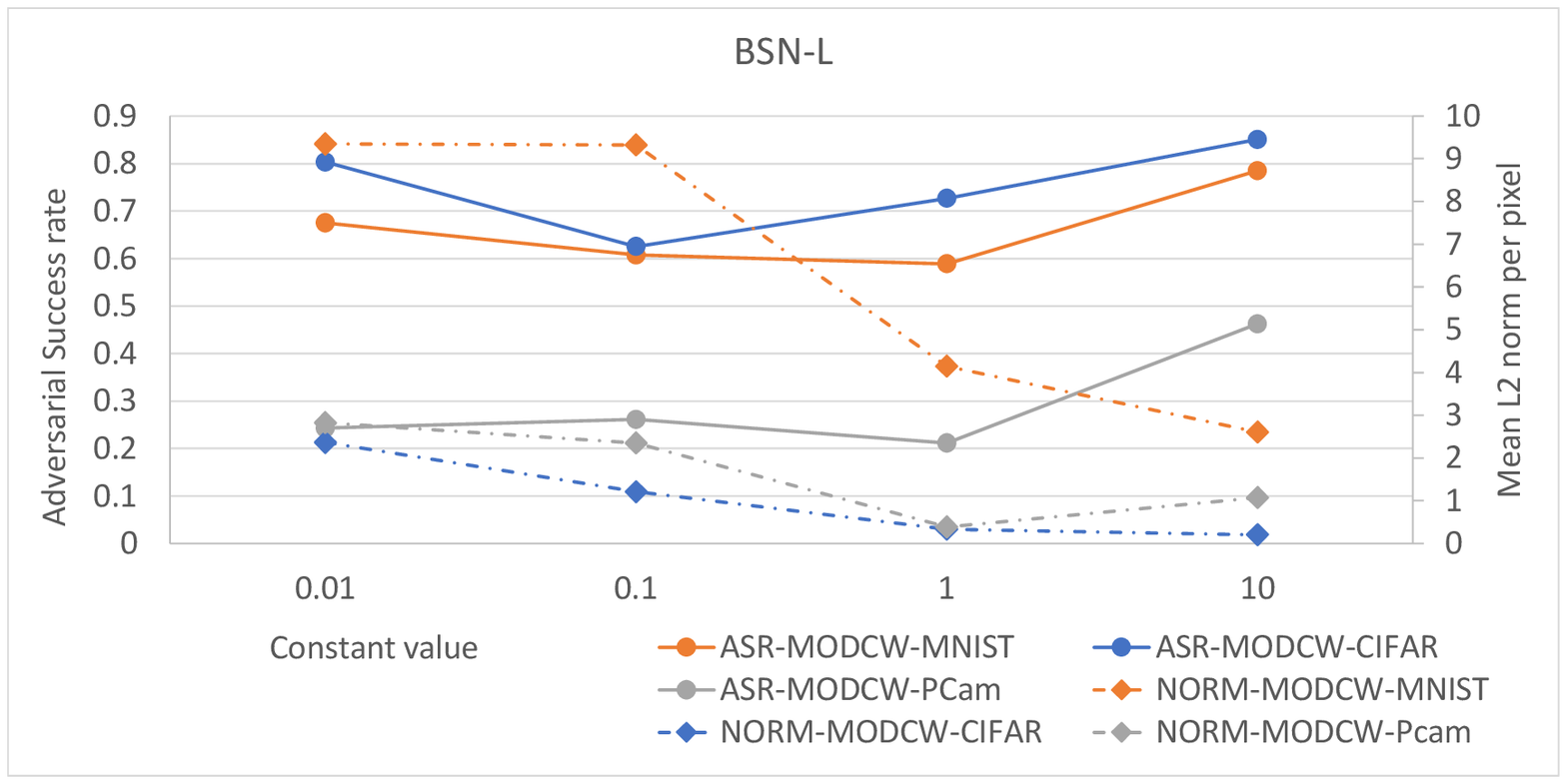}
        \caption{ModCWL2 attack}
        \label{cvaluetrendb}
    \end{subfigure}
    \caption{ASR (solid) and Mean $L_2$ norms per pixel (dashed) against increasing constant value $c$ for BSN-L model on MNIST (orange), CIFAR-10 (blue) and PCam (Grey). Best viewed in colour.}
    \label{cvaluetrend}
\end{figure*}
Interestingly, we observed for the ModCWL2 attack that increasing the value of $c$ decreases the resultant distortion while also increasing ASR in general (see Figure~\ref{cvaluetrend}).
This behaviour is consistent across the different stochastic model variants, as shown in Tables~\ref{tab1:cwattackfullstoc} and \ref{tab1:cwattackl2normsfullstoc}. We postulate that as $c$ increases, the weight of the third term in Equation 3 decreases, thereby reducing the amount of noise introduced during the optimisation process. The high start scale for mean L2 distortion in Figure~\ref{cvaluetrendb}, as compared to \ref{cvaluetrenda}, confirms this.

\subsection{Impact of Stochastic ANNs' Inference Policy on ASR}

In our work, we assumed that the stochastic ANNs we investigated employed an average inference policy, where the predicted network logits was an average of 10 independent forward passes of the same sample. However, we also questioned how ASR varies when the inference policy changes to non-average (i.e. single). The adoption of such a policy improves the run-time complexity.



\begin{figure}[htb]
    \centering
    \begin{subfigure}[t]{0.5\linewidth}
        \centering
        \includegraphics[trim={0cm 0cm 1.0cm 0.6cm},clip, width=1\linewidth]{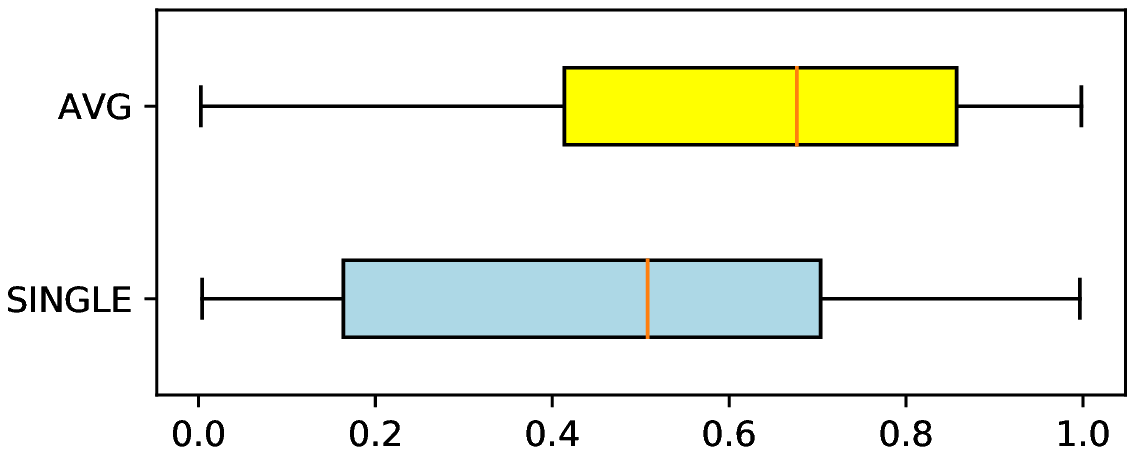}
        \caption{ASR across the different attacks and stochastic models}
    \end{subfigure}%
    ~ 
    \begin{subfigure}[t]{0.5\linewidth}
        \centering
        \includegraphics[trim={0cm 0cm 1.0cm 0.6cm},clip,width=1\linewidth]{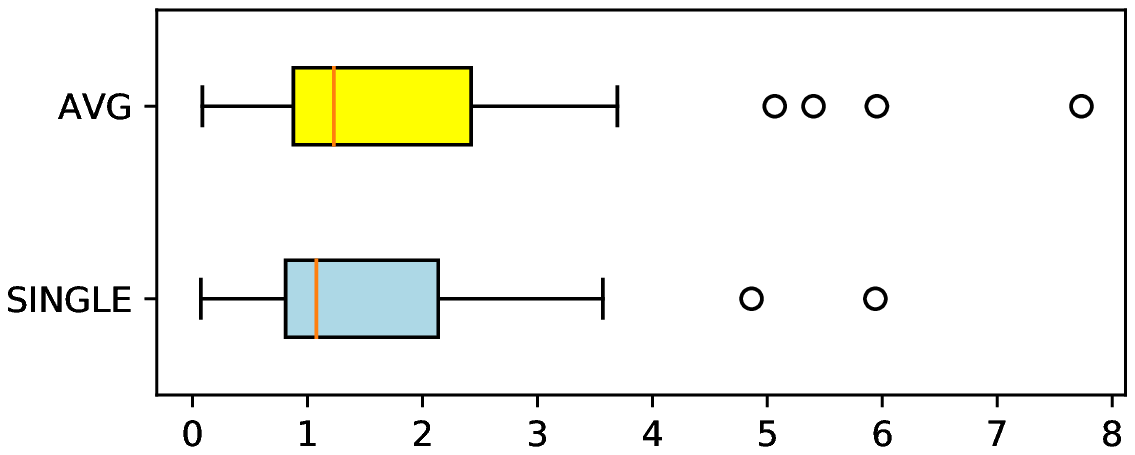}
        \caption{Mean $L_2$ norms per pixel across the different attacks and stochastic models}
    \end{subfigure}
    \caption{Comparison between single and average inference policy in stochastic ANNs. We considered all attacks across the stochastic ANNs, across the datasets when plotting the boxplot.}
    \label{singlevsavg}
\end{figure}

As illustrated in Figure~\ref{singlevsavg}, adopting an average policy yields a higher ASR against stochastic ANNs. The average policy yields a higher median and lower variance than the single policy in terms of ASR. 
This shows the difficulty of attacking stochastic ANNs due to the noise induced by the models.
Hence, for an attacker to attain a more effective adversarial sample, it is better to attack the model based on the average predictions per sample.

\subsection{Implementation Details of Surrogate-based Attacks}

\subsubsection{Training our Surrogate Models}
We describe the hyperparameters we used for training our surrogate models. For the case of a ResNet18 surrogate, we adopted an outer surrogate training epoch, $\rho$, of 5 (i.e. performing dataset augmentation 5 times), and an inner surrogate training epoch of 10 (i.e. classical training of the surrogate model from scratch). Our learning rate was 0.01 and our step size for the Jacobian dataset augmentation, $\lambda$, was 0.1.

For the case of a stochastic surrogate (i.e. BSN-L), we adopted an outer surrogate training epoch of 5, and an inner surrogate training epoch of 20. Our learning rate and $\lambda$ remained the same. Details with regards to the hyperparameters can be found in \cite{papernot2017practical}.

Another parameter defined in \cite{papernot2017practical} was $\tau$, which is the iteration period that controls how often $\lambda$ flips in sign. More specifically, $\lambda$ is defined as such:

\begin{align}
    \lambda_\rho = \lambda * (-1)^{\floor*{\frac{\rho}{\tau}}}
\end{align}{}

$\tau$ was defined as 2 in our experiments.

\subsubsection{Evaluation}
When performing surrogate-based attacks using a stochastic surrogate model, we ensured that the resultant adversarial samples were of reasonable quality before launching them against our target model. We did this by calculating the mean ASR of the adversarial sample, based on 100 forward passes. We consider it as a valid adversarial sample only if the mean ASR was above 0.75. Also, we evaluated the ASR against the targeted model on a fixed number of adversarial samples in our experiments. For instance, if we defined the value to be 100, we ensured that we will always evaluate the ASR against the targeted model with 100 valid adversarial samples in all our experiments. The same was also performed in our transferability experiments.

\subsection{Variance Mimicking}

We provide a pseudo-code of our proposed Variance Mimicking method to train stochastic surrogates, as described in Section 3.4 of our main text, in Algorithm~\ref{vm}. This simply replaces the training of the surrogate model in line 6 of Algorithm 1 in \cite{papernot2017practical}. The other steps remain the same and $m$ was defined as 10.

\begin{algorithm}[htb]
\SetAlgoLined
\KwInput{$O$, $F$, $D$, $m$, $max_\rho$}
\For{$\rho \leftarrow 0$ \KwTo $max_{\rho}$}{
    \For{$B \in D$}{
        \tcp{$d_{B}$ is a data mini-batch}
        \tcp{train $F$ from hard label predictions from $O$}
        $F$.train\_step($d_B$, $O(d_B)$)\\
        $\sigma_{B}^2 \leftarrow get\_variance(d_{B})$\\
        \For{$i \leftarrow 0$ \KwTo $m$}{
            $d_{perturb} \leftarrow d_B + \mathcal{N}(0, \sigma_{B}^2)$\\
            $F$.train\_step($d_{perturb}$, $O(d_{perturb})$)
        }
    }
}
\caption{\textbf{Variance Mimicking -} given oracle $O$, surrogate model $F$, number of training epochs $max_\rho$, number of perturbations $m$ and some surrogate training dataset $D$.}
\label{vm}
\end{algorithm}

\subsection{Adversarial Images of Various Attacks Across Datasets and Models}

\begin{figure*}[htb]
    \centering
    \includegraphics[clip,width=0.9\linewidth]{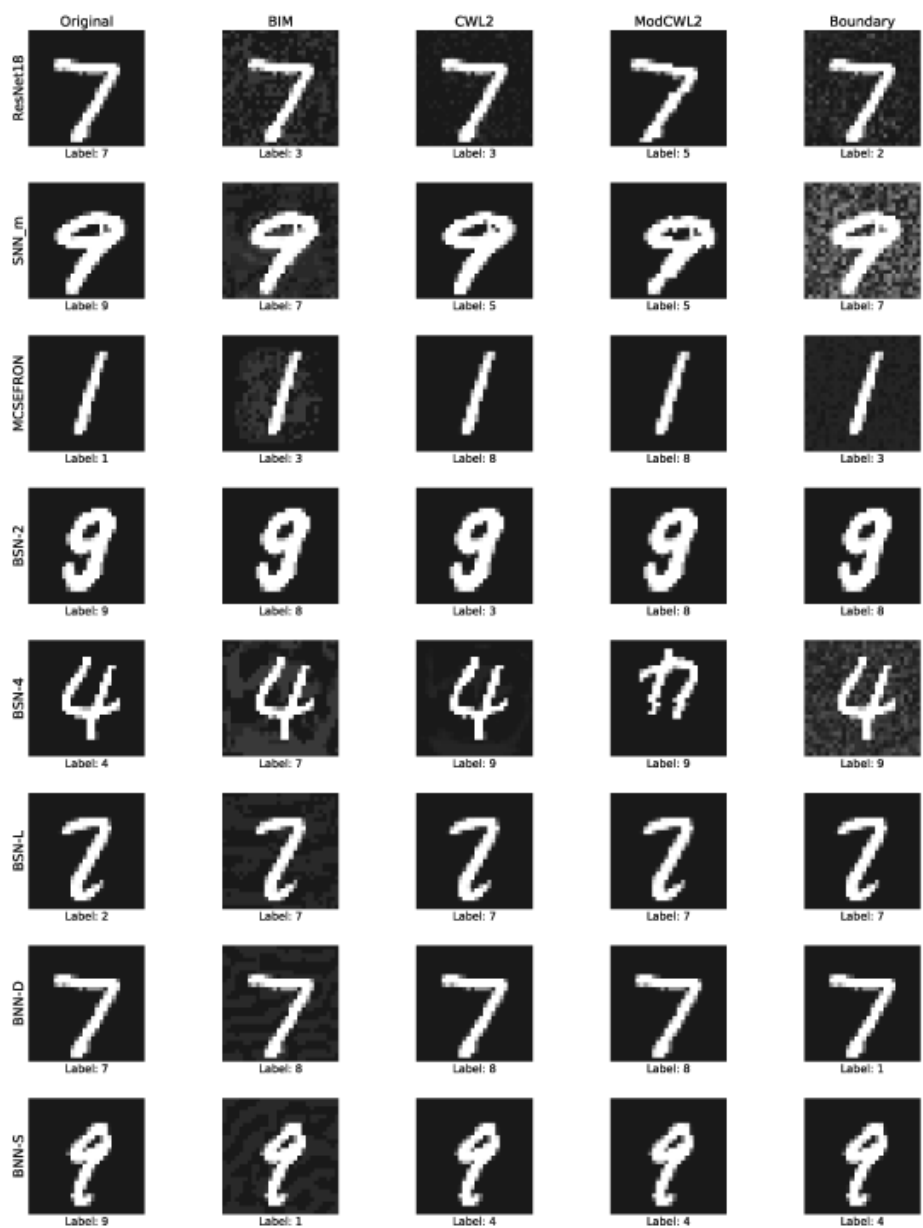}
    \caption{Sample adversarial images from the different attack methods against the different variants of neural networks on the MNIST dataset.}
    \label{mnistwall}
\end{figure*}

Here, we illustrate some sample adversarial images, compared to their original counterparts, for the different attacks and also against the various targeted models in Figure~\ref{mnistwall}. We only show images from the MNIST dataset as they are the most relatable among the three datasets. The first column shows the original image that was used to construct the respective adversarial samples. The subsequent columns represent the different attack methods while the rows represent the different model variants. Below each image, there is a label that indicates what the predicted label was based on the corresponding model.

\end{document}